\documentclass{article}

\usepackage[nonatbib,final]{nips_2018}

\usepackage[utf8]{inputenc} 
\usepackage[T1]{fontenc}    
\usepackage{hyperref}       
\usepackage{url}            
\usepackage{booktabs}       
\usepackage{amsfonts}       
\usepackage{nicefrac}       
\usepackage{microtype}      
\usepackage{graphicx}
\usepackage{tabularx,ragged2e}
\usepackage{amsmath,amssymb} 
\usepackage{calligra}
\usepackage{bbm}
\usepackage{xcolor}
\usepackage{amsmath}
\usepackage{pifont}
\usepackage{array}

\usepackage{algpseudocode,algorithm,algorithmicx}
\algnewcommand{\LineComment}[1]{\State \(\triangleright\) #1}
\algdef{SE}[DOWHILE]{Do}{doWhile}{\algorithmicdo}[1]{\algorithmicwhile\ #1}%

\newcommand{\cmark}{\text{\ding{51}}}

\newcommand{\myparagraph}[1]{\textbf{#1.}}
\newcommand{\mysection}[1]{\vspace{-0.5em}\section{#1}\vspace{-0.5em}}
\newcommand{\mysubsection}[1]{\vspace{-0.25em}\subsection{#1}\vspace{-0.25em}}
\newcommand{\mysubsubsection}[1]{\vspace{-0.25em}\subsubsection{#1}\vspace{-0.25em}}

\DeclareMathOperator*{\argmax}{arg\,max}
\title{Speaker-Follower Models for \\Vision-and-Language Navigation}

\author{
Daniel Fried\thanks{,$^{**}$: Authors contributed equally}$~~^{1}$, Ronghang Hu$^{*1}$, Volkan Cirik$^{*2}$, Anna Rohrbach$^1$, Jacob Andreas$^1$,\\ {\bf Louis-Philippe Morency$^{2}$, Taylor Berg-Kirkpatrick$^{2}$, Kate Saenko$^{3}$,} \\{\bf Dan Klein$^{**1}$, Trevor Darrell$^{**1}$}\\
$^1$University of California, Berkeley $\quad$ $^2$Carnegie Mellon University $\quad$ $^3$Boston University \\
  \vspace{-2em}
}

\begin{document}

\maketitle
\setcounter{footnote}{0}

\begin{abstract}
Navigation guided by natural language instructions presents a challenging reasoning problem for instruction followers.
Natural language instructions typically identify only a few high-level decisions and landmarks rather than complete low-level motor behaviors; much of the missing information must be inferred based on perceptual context.
In machine learning settings, this is doubly challenging: it is difficult to collect enough annotated data to enable learning of this reasoning process from scratch, and also difficult to implement the reasoning process using generic sequence models. Here we describe an approach to vision-and-language navigation that addresses both these issues with an embedded \emph{speaker model}. We use this speaker model to (1) synthesize new instructions for data augmentation and to (2) implement pragmatic reasoning, which evaluates how well candidate action sequences explain an instruction. Both steps are supported by a panoramic action space that reflects the granularity of human-generated instructions. Experiments show that all three components of this approach---speaker-driven data augmentation, pragmatic reasoning and panoramic action space---dramatically improve the performance of a baseline instruction follower, more than doubling the success rate over the best existing approach on a standard benchmark.

\end{abstract}
\vspace{-1em}\mysection{Introduction}

\begin{figure}[t]
\centering
\includegraphics[width=\linewidth]{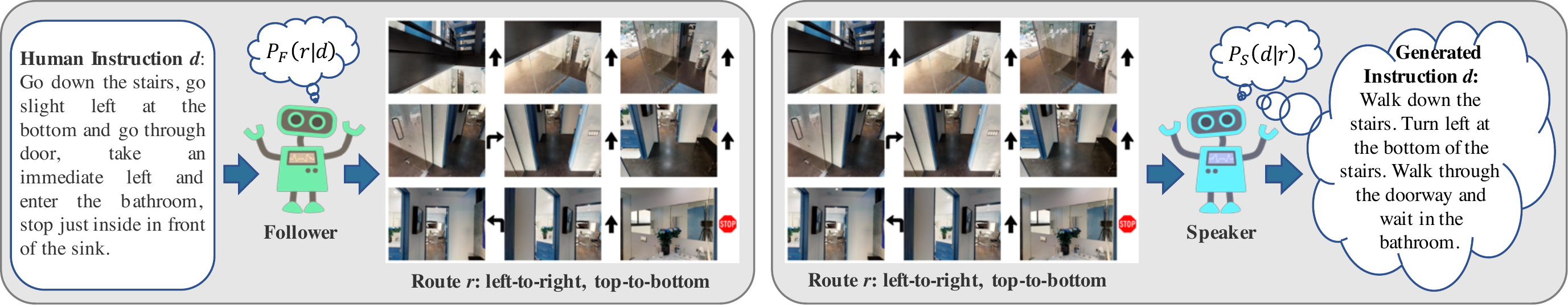}
\vspace{-1.5em}
\caption{The task of vision-and-language navigation is to perform a sequence of actions (navigate through the environment) according to human natural language instructions. Our approach consists of an instruction \emph{follower} model (left) and a \emph{speaker} model (right).} 
\label{fig:teaser}
\vspace{-1.5em}
\end{figure}

In the vision-and-language navigation task \cite{anderson2018cvpr}, an agent is placed in a realistic environment, and provided a natural language instruction such as ``\textit{Go down the stairs, go
slight left at the bottom and go through door, take an immediate left and enter the bathroom, stop just inside in front of the sink}''.
The agent must follow this instruction to navigate from its starting location to a goal location, as shown in Figure \ref{fig:teaser} (left). 
To accomplish this task the agent must learn to relate the language instructions to the visual environment.
Moreover, it should be able to carry out new instructions in unseen environments.

Even simple navigation tasks require nontrivial \emph{reasoning}: the agent must resolve ambiguous references to landmarks, perform a counterfactual evaluation of alternative routes, and identify incompletely specified destinations. 
While a number of approaches \cite{Mei16Instructions,misra2017instructions,wang2018look} have been proposed for the various navigation benchmarks, they generally employ a single model that learns to map directly from instructions to actions from a limited corpus of annotated trajectories.

In this paper we treat the vision-and-language navigation task as a trajectory search problem, where the agent needs to find (based on the instruction) the best trajectory in the environment to navigate from the start location to the goal location. Our model involves an instruction interpretation (\emph{follower}) module, mapping instructions to action sequences; and an instruction generation (\emph{speaker}) module, mapping action sequences to instructions (Figure~\ref{fig:teaser}), both implemented with standard sequence-to-sequence architectures. 
The speaker learns to give textual instructions for visual routes, while the follower learns to follow routes (predict navigation actions) for provided textual instructions. 
Though explicit probabilistic reasoning combining speaker and follower agents is a staple of the literature on computational pragmatics \cite{frank2012predicting}, application of these models has largely been limited to extremely simple decision-making tasks like single forced choices. 

\begin{figure}[t]
\centering
\includegraphics[width=\linewidth]{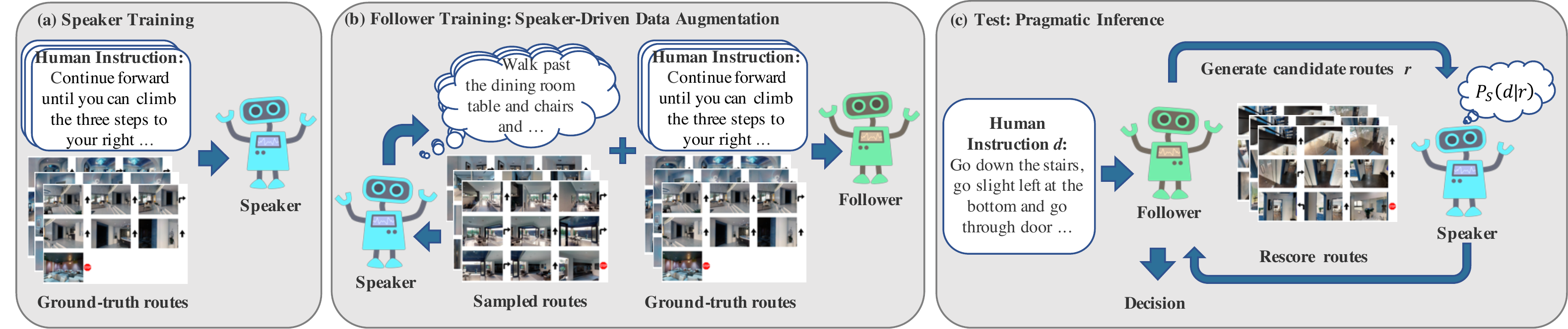}
\vspace{-1.5em}
\caption{Our approach combines an instruction \emph{follower} model and a \emph{speaker} model. (a) The speaker model is trained on the ground-truth routes with human-generated descriptions; (b) it provides the follower with additional synthetic instruction data to bootstrap training; (c) it also helps the follower interpret ambiguous instructions and choose the best route during inference. See Sec.~\ref{sec:method} for details.}
\label{fig:method_overview}
\vspace{-1.5em}
\end{figure}

\begin{figure}[t]
\centering
\includegraphics[width=\linewidth]{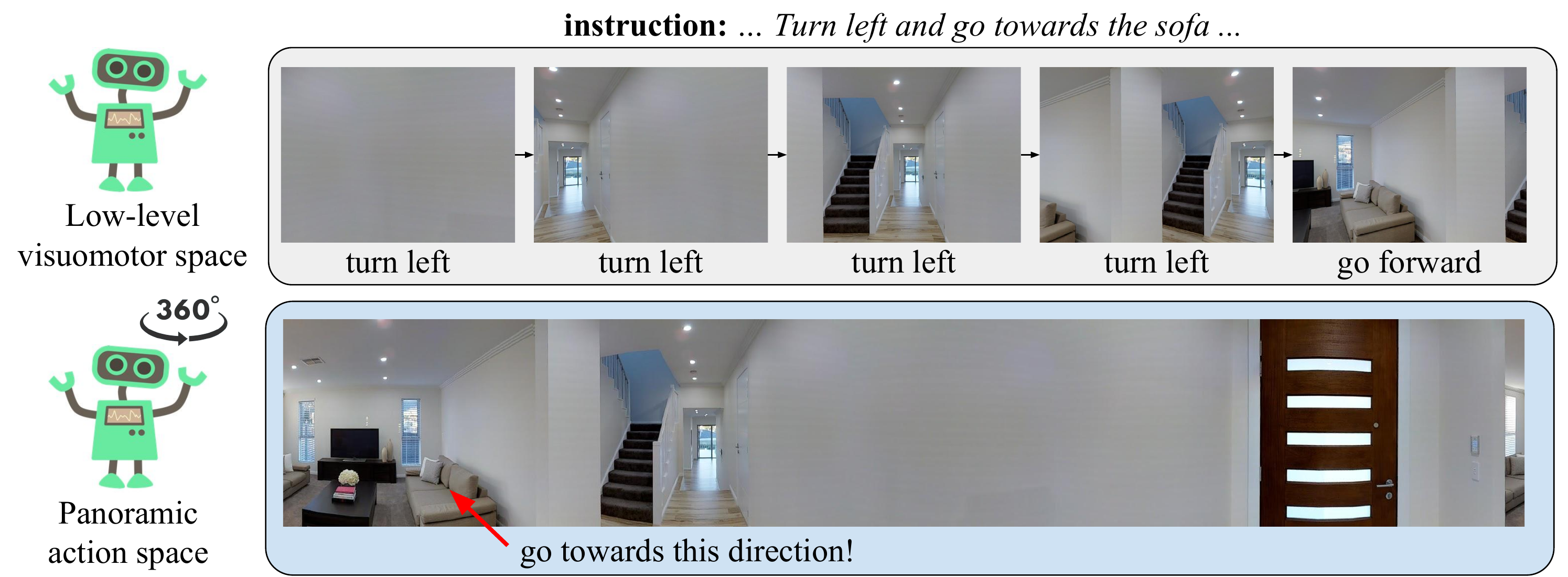}
\vspace{-1.5em}
\caption{Compared with low-level visuomotor space, our panoramic action space (Sec. \ref{sec:method_panoramic}) allows the agents to have a complete perception of the scene, and to directly perform high-level actions.}
\label{fig:method_panoramic}
\vspace{-1.5em}
\end{figure}

We incorporate the speaker both at training time and at test time, where it works together with the learned instruction follower model to solve the navigation task (see Figure~\ref{fig:method_overview} for an overview of our approach). At training time, we perform speaker-driven data augmentation where the speaker helps the follower by synthesizing additional route-instruction pairs to expand the limited training data. 
At test time, the follower improves its chances of success by looking ahead at possible future routes and pragmatically choosing the best route by scoring them according to the probability that the speaker would generate the correct instruction for each route.  This procedure, using the external speaker model, improves upon planning using only the follower model.
We construct both the speaker and the follower on top of a panoramic action space that efficiently encodes high-level behavior, moving directly between adjacent locations rather than making low-level visuomotor decisions like the number of degrees to rotate (see Figure~\ref{fig:method_panoramic}).

To summarize our contributions: We propose a novel approach to vision-and-language navigation incorporating a visually grounded speaker--follower model, and introduce a panoramic representation to efficiently represent high-level actions. We evaluate this speaker--follower model on the Room-to-Room (R2R) dataset \cite{anderson2018cvpr}, and show that each component in our model improves performance at the instruction following task. Our model obtains a final success rate of 53.5\% on the unseen test environment, an absolute improvement of 30\% over existing approaches. Our code and data are available at \url{http://ronghanghu.com/speaker_follower}.

\mysection{Related Work}
\label{sec:related_work}

\myparagraph{Natural language instruction following}
Systems that learn to carry out natural language instructions in an interactive environment include approaches based on intermediate structured and executable representations of language \cite{tellex2011understanding,chen2012grounded,artzi2013instructions,long2016simpler,guu2017bridging} and approaches that map directly from language and world state observations to actions \cite{Branavan09PG,Andreas15Instructions,Mei16Instructions,misra2017instructions}.
The embodied vision-and-language navigation task studied in this paper \cite{anderson2018cvpr} differs from past situated instruction following tasks by introducing rich visual contexts. 
Recent work \cite{wang2018look} has applied techniques from model-based and model-free reinforcement learning \cite{weber2017imagination} to the vision-and-language navigation problem. Specifically, an environment model is used to predict a representation of the state resulting from an action, and planning is performed with respect to this environment model. Our work differs from this prior work 
by reasoning not just about state transitions, but also about the relationship between states and the language that describes them---specifically, using an external speaker model to predict how well a given sequence of states explains an instruction.

\myparagraph{Pragmatic language understanding}
A long line of work in linguistics, natural language processing, and cognitive science has studied \emph{pragmatics}: how linguistic meaning is affected by context and communicative goals \cite{Grice75}. Our work here makes use of the Rational Speech Acts framework \cite{frank2012predicting,goodman2013knowledge}, which models the interaction between speakers and listeners as a process where each agent reasons probabilistically about the other to maximize the chances of successful communicative outcomes. 
This framework has been applied to model human use of language \cite{Frank09PragmaticExperiments}, and to improve the performance of systems that generate \cite{andreas2016reasoning,mao2016generation,vedantam2017context,cohn2018pragmatically} and interpret \cite{yu2017joint,luo17cvpr,vasudevan2018referring} referential language.
Similar modeling tools have recently been applied to generation and interpretation of language about sequential decision-making \cite{fried2017unified}.
The present work makes use of a pragmatic instruction follower in the same spirit. Here, however, we integrate this with a more complex visual pipeline and use it not only at inference time but also at \emph{training} time to improve the quality of a base listener model.

\myparagraph{Semi- and self-supervision}
The semi-supervised approach we use is related to model bootstrapping techniques such as self-training \cite{scudder1965probability,mcclosky2006effective} and co-training \cite{blum1998combining} at a high-level.
Recent work has used monolingual corpora to improve the performance of neural machine translation models structurally similar to the sequence-to-sequence models we use \cite{gulcehre2015using,he2016deep,sennrich2016mt}.
In a grounded navigation context, \cite{hermann2017grounded} use a word-prediction task as training time supervision for a reinforcement learning agent. 
The approach most relevant to our work is the SEQ4 model \cite{kovcisky2016semisupervised}, which applies semi-supervision to a navigation task by sampling new environments and maps (in synthetic domains without vision), and training an autoencoder to reconstruct routes, using language as a latent variable.
The approach used here is much simpler, as it does not require constructing a differentiable surrogate to the decoding objective.

Semi-supervised data augmentation has also been widely used in computer vision tasks.
In Data Distillation \cite{radosavovic2017data}, additional annotation for object and key-point detection is obtained by ensembling and refining a pretrained model's prediction on unannotated images.
Self-play in adversarial groups of agents is common in multi-agent reinforcement learning  \cite{silver2017mastering,sukhbaatar2017intrinsic}.
In actor-critic approaches \cite{sutton1998reinforcement,sutton2000policy} in reinforcement learning, a critic learns the value of a state and is used to provide supervision to the actor's policy during training.
In this work, we use a speaker to synthesize additional navigation instructions on unlabeled new routes, and use this synthetic data from the speaker to train the follower.

\myparagraph{Grounding language in vision}
Existing work in visual grounding \cite{plummer15iccv,mao2016generation,hu16cvpr,rohrbach16eccv,nagaraja16eccv} has addressed the problem of \textit{passively} perceiving a static image and mapping a referential expression to a bounding box \cite{plummer15iccv,mao2016generation,hu16cvpr} or a segmentation mask \cite{hu16eccv,liu2017recurrent,yu2018mattnet}, exploring various techniques including proposal generation \cite{chen2017query} and relationship handling \cite{wang16eccv,nagaraja16eccv,hu17cvpr,cirik2018using}. In our work, the vision-and-language navigation task requires the agent to \textit{actively} interact with the environment to find a path to the goal following the natural language instruction. This can be seen as a grounding problem in linguistics where the language instruction is grounded into a trajectory in the environment but requires more reasoning and planning skills than referential expression grounding.

\mysection{Instruction Following with Speaker-Follower Models}
\label{sec:method}

To address the task of following natural language instructions, we rely on two models: an instruction-\emph{follower} model of the kind considered in previous work, and a \emph{speaker} model---a learned instruction generator that models how humans describe routes in navigation tasks.

Specifically, we base our follower model on the sequence-to-sequence model \cite{anderson2018cvpr}, computing a distribution $P_F(r \mid d)$ over routes $r$ (state and action sequences) given route descriptions $d$. 
The follower encodes the sequence of words in the route description with an LSTM \cite{hochreiter1997long}, and outputs route actions sequentially, using an attention mechanism \cite{bahdanau2014neural} over the description. Our speaker model is symmetric, producing a distribution $P_S(d \mid r)$ by encoding the sequence of visual observations and actions in the route using an LSTM, and then outputting an instruction word-by-word with an LSTM decoder using attention over the encoded input route (Figure \ref{fig:teaser}).

We combine these two base models into a speaker-follower system, where the speaker supports the follower both at training time and at test time. An overview of our approach is presented in Figure~\ref{fig:method_overview}. First, we train a speaker model on the available ground-truth navigation routes and instructions.
(Figure~\ref{fig:method_overview} (a)).
Before training the follower, the speaker produces synthetic navigation instructions for novel sampled routes in the training environments, which are then used as additional supervision for the follower, as described in Sec. \ref{sec:method_augmentation} (Figure~\ref{fig:method_overview} (b)). At follower test time, the follower generates possible routes as interpretations of a given instruction and starting context, and the speaker pragmatically ranks these, choosing one that provides a good explanation of the instruction in context (Sec.\ \ref{sec:method_pragmatics} and Figure~\ref{fig:method_overview} (c)). Both follower and speaker are supported by the panoramic action space in Sec. \ref{sec:method_panoramic} that reflects the high-level granularity of the navigational instructions (Figure \ref{fig:method_panoramic}).

\mysubsection{Speaker-Driven Data Augmentation}
\label{sec:method_augmentation}

The training data only covers a limited number of navigation instruction and route pairs, $\mathcal{D} = (d_1, r_1) \ldots (d_N, r_N)$. To allow the agent to generalize better to new routes, we use the speaker to generate synthetic instructions on sampled new routes in the training environments. 
To create a synthetic training set, we sample a collection of $M$ routes $\hat r_1, \ldots, \hat r_M$ through the training environments, using the same shortest-path approach used to generate the routes in the original training set \cite{anderson2018cvpr}. We then generate a human-like textual instruction $\hat d_k$ for each instruction $\hat r_k$ by performing greedy prediction in the speaker model to approximate $\hat d_k = \argmax_d P_S(d \mid \hat r_k)$.

These $M$ synthetic navigation routes and instructions $\mathcal{S} = (\hat d_1, \hat r_1), \ldots, (\hat s_M, \hat r_M)$ are combined with the original training data $\mathcal{D}$ into an augmented training set $\mathcal{S} \cup \mathcal{D}$ (Figure~\ref{fig:method_overview}(b)). During training, the follower is first trained on this augmented training set, and then further fine-tuned on the original training set $\mathcal{D}$. This speaker-driven data augmentation aims to overcome data scarcity issue, allowing the follower to learn how to navigate on new routes following the synthetic instructions.

\mysubsection{Speaker-Driven Route Selection}
\label{sec:method_pragmatics}

We use the base speaker ($P_S$) and follower ($P_F$) models described above to define a \emph{pragmatic follower} model. Drawing on the Rational Speech Acts framework \cite{frank2012predicting,goodman2013knowledge}, a pragmatic follower model should choose a route $r$ through the environment that has high probability of having caused the speaker model to produce the given description $d$: 
$\argmax_{r} P_{S}(d \mid r)$ (corresponding to a rational Bayesian follower with a uniform prior over routes).
Such a follower chooses a route that provides a good explanation of the observed description, allowing counterfactual reasoning about instructions, or using global context to correct errors in the follower's path, which we call \textit{pragmatic inference}.

Given the sequence-to-sequence models that we use, exactly solving the maximization problem above is infeasible; and may not even be desirable, as these models are trained discriminatively and may be poorly calibrated for inputs dissimilar to those seen during training.
Following previous work on pragmatic language generation and interpretation \cite{smith2013learning,andreas2016reasoning,monroe2017colors,fried2017unified}, we use a rescoring procedure: produce candidate route interpretations for a given instruction using the base follower model, and then rescore these routes using the base speaker model (Figure~\ref{fig:method_overview}(c)).

Our pragmatic follower produces a route for a given instruction by obtaining $K$ candidate paths from the base follower using a search procedure described below, then chooses the highest scoring path under a combination of the follower and speaker model probabilities:
\begin{equation}
\label{eqn:pragmatics}
\argmax_{r \in R(d)} P_{S}(d \mid r)^\lambda \cdot P_{F}(r \mid d)^{(1 - \lambda)}
\end{equation}
where $\lambda$ is a hyper-parameter in the range $[0, 1]$ which we tune on validation data to maximize the accuracy of the follower.\footnote{In practice, we found best performance with values of $\lambda$ close to 1, relying mostly on the score of the speaker to select routes. Using only the speaker score (which corresponds to the standard RSA pragmatic follower) did not substantially reduce performance compared to using a combination with the follower score, and both improved substantially upon using only the follower score (corresponding to the base follower).}

\myparagraph{Candidate route generation}
To generate candidate routes from the base follower model, we perform a search procedure where candidate routes are produced incrementally, action-by-action, and scored using the probabilities given by $P_F$. 
Standard beam search in sequence-to-sequence models (e.g. \cite{sutskever2014sequence}) forces partial routes to compete based on the number of actions taken.
We obtain better performance by instead using a \emph{state-factored} search procedure, where partial output sequences compete at the level of states in the environment, where each state consists of the agent's location and discretized heading, keeping only the highest-scoring path found so far to each state. 
At a high-level, this search procedure resembles graph search with a closed list, but since action probabilities are non-stationary (potentially depend on the entire sequence of actions taken in the route), it is only approximate, and so we allow re-expanding states if a higher-scoring route to that state is found.

At each point in our state-factored search for searching and generating candidates in the follower model, we store the highest-probability route (as scored by the follower model) found so far to each state. States contain the follower's discrete location and heading (direction it is facing) in the environment, and whether the route has been completed (had the \textsc{Stop} action predicted). The highest-scoring route, which has not yet been expanded (had successors produced), is selected and expanded using each possible action from the state, producing routes to the neighboring states. For each of these routes $r$ with final state $s$, if $s$ has not yet been reached by the search, or if $r$ is higher-scoring under the model than the current best path to $s$, $r$ is stored as the best route to $s$. We continue the search procedure until $K$ routes ending in distinct states have predicted the \textsc{Stop} action, or there are no remaining unexpanded routes. 
See Sec.~B in the supplementary material for pseudocode.

Since route scores are products of conditional probabilities, route scores are non-increasing, and so this search procedure generates routes that do not pass through the same state twice---which we found to improve accuracy both for the base follower model and the pragmatic rescoring procedure, since instructions typically describe acyclic routes.

We generate up to $K=40$ candidate routes for each instruction using this procedure, and rescore using Eq.~\ref{eqn:pragmatics}. In addition to enabling pragmatic inference, this state-factored search procedure improves the performance of the follower model on its own (taking the candidate route with highest score under the follower model), when compared to standard greedy search (see Sec.~C and Figure~C.2 of the supplementary material for details).

\mysubsection{Panoramic Action Space}
\label{sec:method_panoramic}

The sequence-to-sequence agent in \cite{anderson2018cvpr} uses low-level visuomotor control (such as turning left or right by 30 degrees), and only perceives frontal visual sensory input. Such fine-grained visuomotor control and restricted visual signal introduce challenges for instruction following. For example in Figure \ref{fig:method_panoramic}, to ``turn left and go towards the sofa'', the agent needs to perform a series of turning actions until it sees a sofa in the center of its view, and then perform a ``go forward'' action. This requires strong skills of planning and memorization of visual inputs. While a possible way to address this challenge is to learn a hierarchical policy such as in \cite{das2017embodied}, in our work we directly allow the agent to reason about high-level actions, using a panoramic action space with panoramic representation, converted with built-in mapping from low-level visuomotor control.

As shown in Figure \ref{fig:method_panoramic}, in our panoramic representation, the agent first ``looks around'' and perceives a 360-degree panoramic view of its surrounding scene from its current location, which is discretized into 36 view angles (12 headings $\times$ 3 elevations with 30 degree intervals -- in our implementation). Each view angle $i$ is represented by an encoding vector $v_i$. At each location, the agent can only move towards a few navigable directions (provided by the navigation environment) as other directions can be physically obstructed (e.g. blocked by a table). Here, in our action space the agent only needs to make high-level decisions as to which navigable direction to go to next, with each navigable direction $j$ represented by an encoding vector $u_j$.
The encoding vectors $v_i$ and $u_j$ of each view angle and navigable direction are obtained by concatenating an appearance feature (ConvNet feature extracted from the local image patch around that view angle or direction) and a 4-dimensional orientation feature $[\sin\psi; \cos\psi; \sin\theta; \cos\theta]$, where $\psi$ and $\theta$ are the heading and elevation angles respectively. Also, we introduce a \textsc{Stop} action encoded by $u_0 = \overrightarrow{0}$. The agent can take this \textsc{Stop} action when it decides it has reached the goal location (to end the episode).

To make a decision on which direction to go, the agent first performs one-hop visual attention to look at all of the surrounding view angles, based on its memory vector $h_{t-1}$. The attention weight $\alpha_{t,i}$ of each view angle $i$ is computed as $a_{t,i} = \left(W_1 h_{t-1}\right)^T W_2 v_{t,i}$ and $\alpha_{t,i} = \exp(a_{t,i}) / \sum_i \exp(a_{t,i})$.

The attended feature representation $v_{t,att}=\sum_i \alpha_{t,i} v_{t,i}$ from the panoramic scene is then used as visual-sensory input to the sequence-to-sequence model (replacing the 60-degree frontal appearance vector in \cite{anderson2018cvpr}) to update the agent's memory. Then, a bilinear dot product is used to obtain the probability $p_j$ of each navigable direction $j$: $y_j = \left(W_3 h_t\right)^T W_4 u_j$ and $p_j = \exp(y_j) / \sum_j \exp(y_j)$.

The agent then chooses a navigable direction $u_j$ (with probability $p_j$) to go to the adjacent location along that direction (or $u_0$ to stop and end the episode).
We use a built-in mapping that seamlessly translates our panoramic perception and action into visuomotor control such as turning and moving.

\mysection{Experiments}
\label{sec:experiments}

\mysubsection{Experimental Setup}
\label{sec:exp_setup}

\myparagraph{Dataset}
We use the Room-to-Room (R2R) vision-and-language navigation dataset \cite{anderson2018cvpr} for our experimental evaluation.
In this task, the agent starts at a certain location in an environment and is provided with a human-generated navigation instruction, that describes a path to a goal location. The agent needs to follow the instruction by taking multiple discrete actions (e.g. turning, moving) to navigate to the goal location, and executing a ``stop'' action to end the episode. Note that differently from some robotic navigation settings \cite{pathak2018zero}, here the agent is not provided with a goal image, but must identify from the textual description and environment whether it has reached the goal.

The dataset consists of 7,189 paths sampled from the Matterport3D \cite{Matterport3D} navigation graphs, where each path consists of 5 to 7 discrete viewpoints and the average physical path length is 10m. Each path has three instructions written by humans, giving 21.5k instructions in total, with an average of 29 words per instruction. The dataset is split into training, validation, and test sets. The validation set is split into two parts: \emph{seen}, where routes are sampled from environments seen during training, and \emph{unseen} with environments that are not seen during training. All the test set routes belong to new environments unseen in the training and validation sets.

\myparagraph{Evaluation metrics}
Following previous work on the R2R task, our primary evaluation metrics are navigation error (NE), measuring the average distance between the end-location predicted by the follower agent and the true route's end-location, and success rate (SR), the percentage of predicted end-locations within 3m of the true location. As in previous work, we also report the oracle success rate (OSR), measuring success rate at the closest point to the goal that the follower has visited along the route, allowing the agent to overshoot the goal without being penalized.

\myparagraph{Implementation details}
Following \cite{anderson2018cvpr} and \cite{wang2018look}, we produce visual feature vectors $v$ using the output from the final convolutional layer of a ResNet \cite{he2016deep} trained on the ImageNet \cite{russakovsky2015imagenet} classification dataset. These visual features are fixed, and the ResNet is not updated during training. 
To better generalize to novel words in the vocabulary, we also experiment with using GloVe embeddings \cite{pennington2014glove}, to initialize the word-embedding vectors in the speaker and follower.

In the baseline without using synthetic instructions, we train follower and speaker models using the human-generated instructions for routes present in the training set. The training procedure for the follower model follows \cite{anderson2018cvpr} by training with \emph{student-forcing} (sampling actions from the model during training, and supervising using a shortest-path action to reach the goal state). We use the training split in the R2R dataset to train our speaker model, using standard maximum likelihood training with a cross-entropy loss.

In speaker-driven data augmentation (Sec.~\ref{sec:method_augmentation}), we augment the data used to train the follower model by sampling $178,000$ routes from the training environments. Instructions for these routes are generated using greedy inference in the speaker model (which is trained only on human-produced instructions). The follower model is trained using student-forcing on this augmented data for $50,000$ iterations, and then fine-tuned on the the original human-produced data for $20,000$ iterations. 
For all experiments using pragmatic inference, we use a speaker weight of $\lambda = 0.95$, which we found to produce the best results on both the seen and unseen validation environments.

\mysubsection{Results and Analysis}
\label{sec:exp_quantitative}

We first examine the contribution from each of our model's components on the validation splits.
Then, we compare the performance of our model with previous work on the unseen test split. 

\mysubsubsection{Component Contributions}
\label{sec:exp_ablation}

We begin with a baseline (Row 1 of Table~\ref{tab:results_ablation}), which uses only a follower model with a non-panoramic action space at both training and test time, which is equivalent to the student-forcing model in \cite{anderson2018cvpr}.\footnote{Note that our results for this baseline are slightly higher on val-seen and slightly lower on val-unseen than those reported by \cite{anderson2018cvpr}, due to differences in implementation details and hyper-parameter choices.}

\begin{table}[t]
\small
\begin{center}
\begin{tabular}{rccccccccc}
\toprule
  &  Data & Pragmatic & Panoramic & \multicolumn{3}{c}{Validation-Seen} & \multicolumn{3}{c}{Validation-Unseen} \\
  \cmidrule(l){5-7}\cmidrule(l){8-10}
  \# & Augmentation & Inference & Space & NE $\downarrow$ & SR $\uparrow$ & OSR $\uparrow$ & NE $\downarrow$ & SR $\uparrow$ & OSR $\uparrow$ \\
  \midrule
  1 &   &   &   & 6.08 & 40.3 & 51.6 & 7.90 & 19.9 & 26.1 \\
  \cmidrule(r){1-10}
  2 & $\cmark$ &  &   & 5.05  & 46.8 & 59.9 & 7.30 & 24.6 & 33.2 \\
  3 &   & $\cmark$ &   & 5.23 & 51.5 & 60.8 & 6.62 & 34.5 & 43.1 \\
  4 &   &   & $\cmark$ & 4.86 & 52.1 & 63.3 & 7.07 & 31.2 & 41.3 \\
  \cmidrule(r){1-10}
  5 & $\cmark$ & $\cmark$ &   & 4.28 & 57.2 & 63.9 & 5.75 & 39.3 & 47.0 \\
  6 & $\cmark$ &   & $\cmark$ & 3.36 & 66.4 & 73.8 & 6.62 & 35.5 & 45.0 \\
  7 &   & $\cmark$ & $\cmark$ & 3.88 & 63.3 & 71.0 & 5.24 & 49.5 & 63.4 \\
  \cmidrule(r){1-10}
  8 & $\cmark$ & $\cmark$ & $\cmark$ & \textbf{3.08} & \textbf{70.1} & \textbf{78.3} & \textbf{4.83} & \textbf{54.6} & \textbf{65.2} \\
\bottomrule
\end{tabular}
\end{center}
\caption{Ablations showing the effect of each component in our model. Rows 2-4 show the effects of adding a single component to the baseline system (Row 1); Rows 5-7 show the effects of removing a single component from the full system (Row 8). NE is navigation error (in meters); lower is better. SR and OSR are success rate and oracle success rate (\%); higher is better. See Sec. \ref{sec:exp_ablation} for details.}
\label{tab:results_ablation}
\vspace{-2em}
\end{table}

\myparagraph{Speaker-driven data augmentation}
We first introduce the speaker at training time for data augmentation (Sec. \ref{sec:method_augmentation}). Comparing Row 1 (the baseline follower model trained only with the original training data) against Row 2 (training this model on augmented data) in Table~\ref{tab:results_ablation}, we see that adding the augmented data improves success rate (SR) from 40.3\% to 46.8\% on validation seen and from 19.9\% to 24.6\% on validation unseen, respectively.
This higher relative gain on unseen environments shows that the follower can learn from the speaker-annotated routes to better generalize to new scenes.

Note that given the noise in our augmented data, we fine-tune our model on \textit{the original training data} at the end of training as mentioned in Sec.~\ref{sec:method_augmentation}. We find that increasing the amount of augmented data is helpful in general. For example, when using 25\% of the augmented data, the success rate improves to 22.8\% on validation unseen, while with all the augmented data the success rate reaches 24.6\% on validation unseen, which is a good balance between performance and computation overhead.

\myparagraph{Pragmatic inference}
We then incorporate the speaker at test time for pragmatic inference (Sec. \ref{sec:method_pragmatics}), using the speaker to rescore the route candidates produced by the follower. Adding this technique brings a further improvement in success rate on both environments (compare Row 2, the data-augmented follower without pragmatic inference, to Row 5, adding pragmatic inference).
This shows that when reasoning about navigational directions, large improvements in accuracy can be obtained by scoring how well the route explains the direction using a speaker model. Importantly, when using only the follower model to score candidates produced in search, the success rate is 49.0\% on val-seen and 30.5\% on val-unseen, showing the importance of using the speaker model to choose among candidates (which increases success rates to 57.2\% and 39.3\%, respectively).

\myparagraph{Panoramic action space}
Finally, we replace the visuomotor control space with the panoramic representation (Sec.~\ref{sec:method_panoramic}). Adding this to the previous system (compare Row 5 and Row 8) shows that the new representation leads to a substantially higher success rate, achieving 70.1\% and 54.6\% success rate on validation seen and validation unseen, respectively. This suggests that directly acting in a higher-level representation space makes it easier to accurately carry out instructions.
Our final model (Row 8) has over twice the success rate of the baseline follower in the unseen environments.

\myparagraph{Importance of all components}
Above we have shown the gain from each component, after being added incrementally. Moreover, comparing Rows 2-4 (adding each component independently to the base model) to the baseline (Row 1) shows that each component in isolation provides large improvements in success rates, and decreases the navigation error. Ablating each component (Rows 5-7) from the full model (Row 8) shows that each of them is important for the final performance.

\begin{figure}[b]
\begin{center}

\includegraphics[width=\linewidth]{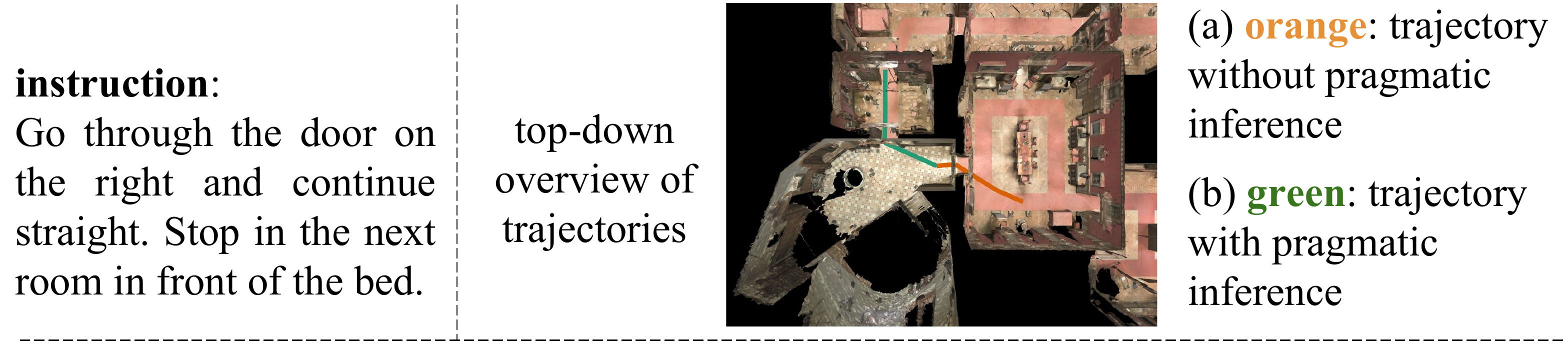}

\begin{tabular}{m{0.8cm}m{12cm}}
\small{Step 1} & \includegraphics[width=\linewidth,trim={5cm 0.75cm 5cm 0.85cm},clip]{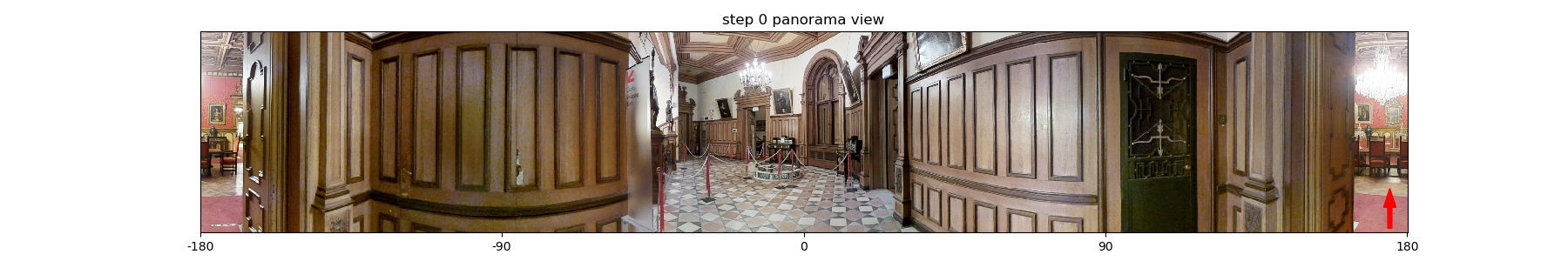} \\
\small{Step 2} & \includegraphics[width=\linewidth,trim={5cm 0.75cm 5cm 0.85cm},clip]{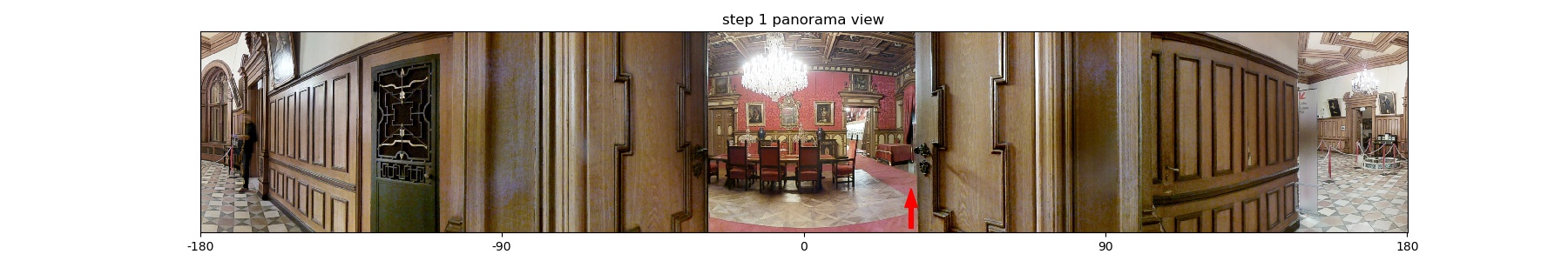} \\
\small{Step 3} & \includegraphics[width=\linewidth,trim={5cm 0.75cm 5cm 0.85cm},clip]{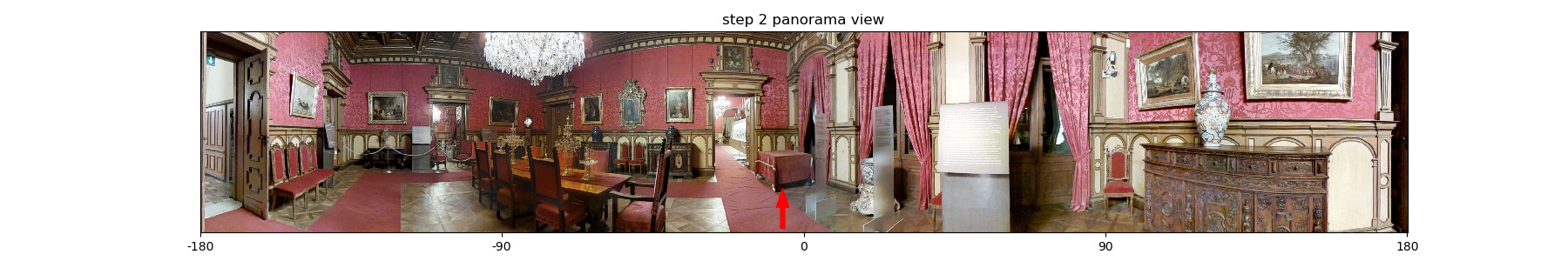} \\
\small{Step 4} & \includegraphics[width=\linewidth,trim={5cm 0.75cm 5cm 0.85cm},clip]{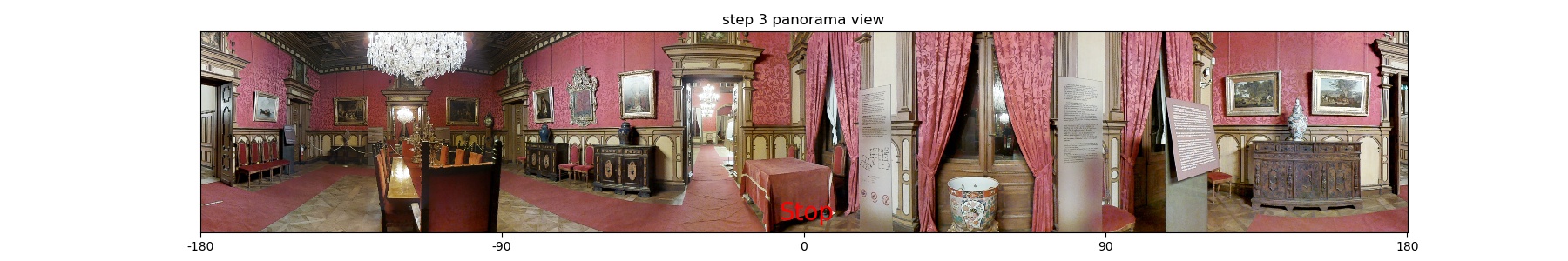} \\
\end{tabular}

{(a) navigation steps \textbf{without pragmatic inference}; \textcolor{red}{red} arrow: direction to go next} \\
\vspace{0.6em}

\begin{tabular}{m{0.8cm}m{12cm}}
\small{Step 1} & \includegraphics[width=\linewidth,trim={5cm 0.75cm 5cm 0.85cm},clip]{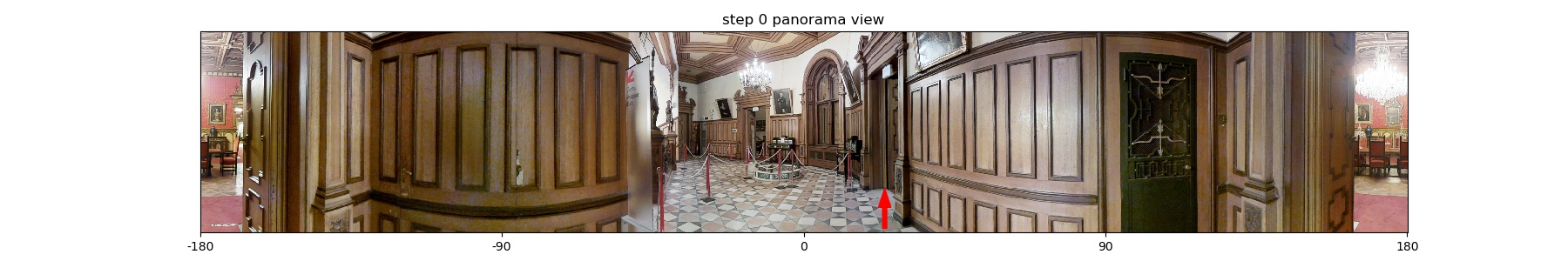} \\
\small{Step 2} & \includegraphics[width=\linewidth,trim={5cm 0.75cm 5cm 0.85cm},clip]{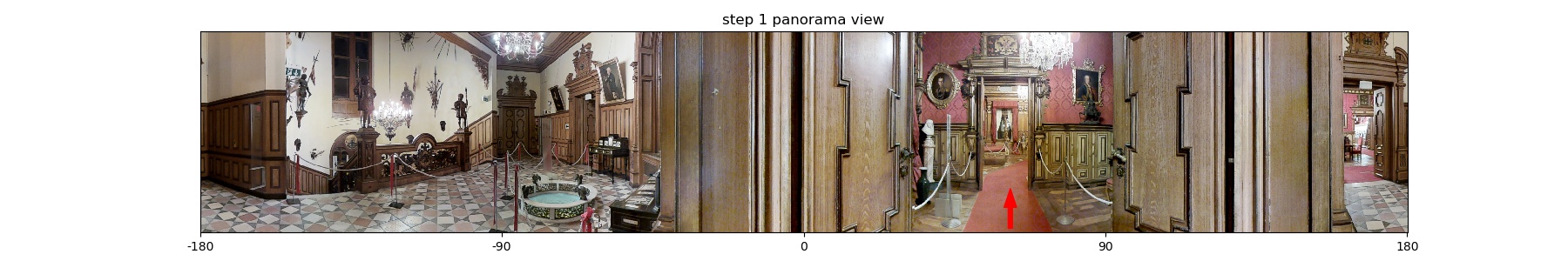} \\
\small{Step 3} & \includegraphics[width=\linewidth,trim={5cm 0.75cm 5cm 0.85cm},clip]{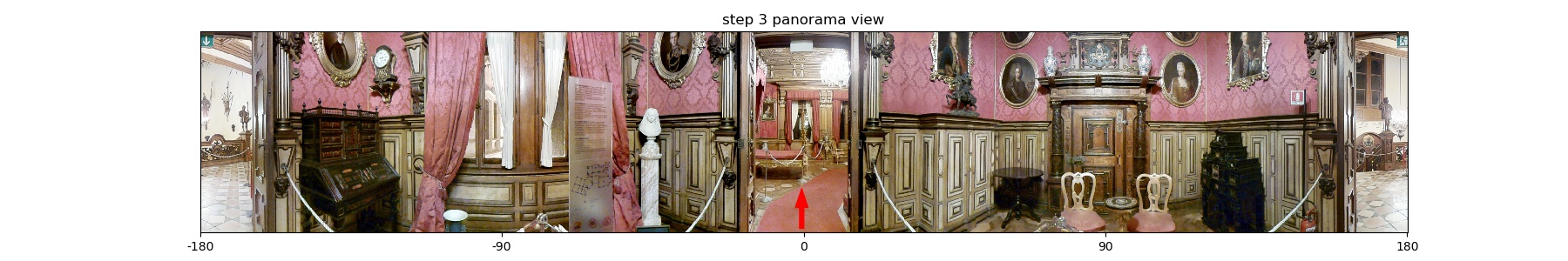} \\
\small{Step 4} & \includegraphics[width=\linewidth,trim={5cm 0.75cm 5cm 0.85cm},clip]{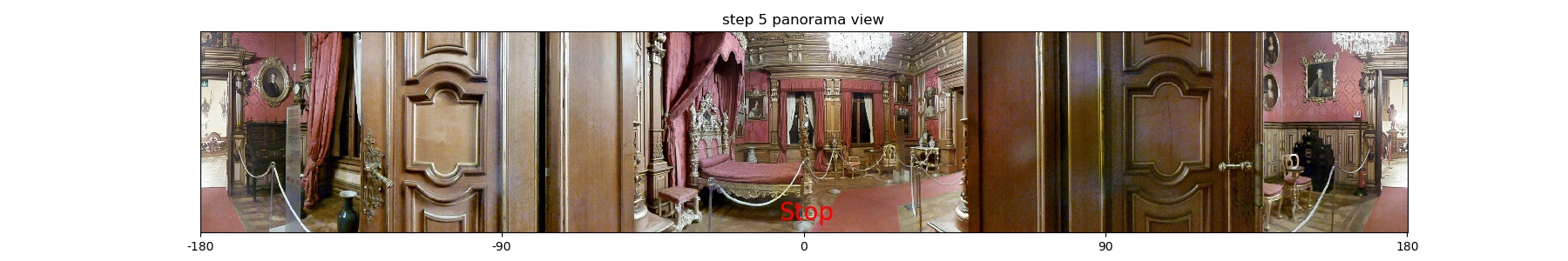} \\
\end{tabular}

{(b) navigation steps \textbf{with pragmatic inference}; \textcolor{red}{red} arrow: direction to go next}
\end{center}
\vspace{-1em}
\caption{Navigation examples on unseen environments with and without pragmatic inference from the speaker model (\textit{best visualized in color}). (a) The follower without pragmatic inference misinterpreted the instruction and went through a wrong door into a room with no bed. It then stopped at a table (which resembles a bed). (b) With the help of a speaker for pragmatic inference, the follower selected the correct route that enters the right door and stopped at the bed.}
\label{fig:qual}
\vspace{-1.5em}
\end{figure}

\myparagraph{Qualitative results}
Here we provide qualitative examples further explaining how our model improves over the baseline (more qualitative results in the supplementary material). The intuition behind the speaker model is that it should help the agent more accurately interpret instructions specifically in ambiguous situations. Figure~\ref{fig:qual} shows how the introduction of a speaker model helps the follower with pragmatic inference.

\mysubsubsection{Comparison to Prior Work}
\label{sec:exp_compare}

We compare the performance of our final model to previous approaches on the R2R held-out splits, including the test split which contains 18 new environments that do not overlap with any training or validation splits, and are only seen once at test time.

The results are shown in Table~\ref{tab:results_sota}. In the table, ``Random'' is randomly picking a direction and going towards that direction for 5 steps. ``Student-forcing'' is the best performing method in \cite{anderson2018cvpr}, using exploration during training of the sequence-to-sequence follower model. ``RPA'' \cite{wang2018look} is a combination of model-based and model-free reinforcement learning (see also Sec. \ref{sec:related_work} for details). ``ours'' shows our performance using the route selected by our pragmatic inference procedure, while ``ours (challenge participation)'' uses a modified inference procedure for submission to the Vision-and-Language Navigation Challenge (See Sec.~E in the supplementary material for details).
Prior work has reported higher performance on the seen rather than unseen environments \cite{anderson2018cvpr,wang2018look}, illustrating the issue of generalizing to new environments. 
Our method more than doubles the success rate of the state-of-the-art RPA approach, and on the test set achieves a final success rate of 53.5\%.
This represents a large reduction in the gap between machine and human performance on this task.

\begin{table}[t]
\scriptsize
\begin{center}
\begin{tabular}{ccccccccccc}
\toprule
& \multicolumn{3}{c}{Validation-Seen} & \multicolumn{3}{c}{Validation-Unseen} & \multicolumn{4}{c}{Test (unseen)} \\
\cmidrule(l){2-4}\cmidrule(l){5-7}\cmidrule(l){8-11}
  Method & NE $\downarrow$ & SR $\uparrow$ & OSR $\uparrow$ & NE $\downarrow$ & SR $\uparrow$& OSR $\uparrow$& NE $\downarrow$& SR $\uparrow$& OSR $\uparrow$ & TL $\downarrow$ \\
\midrule
  Random  & 9.45 & 15.9 & 21.4 & 9.23 & 16.3 & 22.0 & 9.77 & 13.2 & 18.3 & 9.89  \\
  Student-forcing \cite{anderson2018cvpr} & 6.01 & 38.6 & 52.9 & 7.81 & 21.8 & 28.4 & 7.85 & 20.4 & 26.6 & 8.13 \\
  RPA \cite{wang2018look} & 5.56 & 42.9 & 52.6 & 7.65 & 24.6 & 31.8 & 7.53 & 25.3 & 32.5 & 9.15 \\
\midrule
  ours & \textbf{3.08} & \textbf{70.1} & \textbf{78.3} & \textbf{4.83} & \textbf{54.6} & \textbf{65.2} & \textbf{4.87} & \textbf{53.5} & 63.9 & 11.63 \\
  ours (challenge participation)* & -- & -- & -- & -- & -- & -- & \textbf{4.87} & \textbf{53.5} & \textbf{96.0} & 1257.38 \\
  \cmidrule(l){1-11}
  Human & -- & -- & -- & -- & -- & -- & 1.61 & 86.4 & 90.2 & 11.90 \\
\bottomrule
\end{tabular}
\end{center}
\caption{Performance comparison of our method to previous work. NE is navigation error (in meters); lower is better. SR and OSR are success rate and oracle success rate (\%) respectively (higher is better). Trajectory length (TL) on the test set is reported for completeness. \textit{*: When submitting to the Vision-and-Language Navigation Challenge, we modified our search procedure to maintain physical plausibility and to comply with the challenge guidelines. The resulting trajectory has higher oracle success rate while being very long. See Sec.~E in the supplementary material for details.}}
\label{tab:results_sota}
\vspace{-2em}
\end{table}

\mysection{Conclusions}

The language-and-vision navigation task presents a pair of challenging reasoning problems: in language, because agents must interpret instructions in a changing environmental context; and in vision, because of the tight coupling between local perception and long-term decision-making.
The comparatively poor performance of the baseline sequence-to-sequence model for instruction following suggests that more powerful modeling tools are needed to meet these challenges. In this work, we have
introduced such a tool, showing that a follower model for vision-and-language navigation is substantially improved by carefully structuring the action space and integrating an explicit model of a \emph{speaker} that predicts how navigation routes are described. We believe that these results point toward further opportunities for improvements in instruction following by modeling the global structure of navigation behaviors and the pragmatic contexts in which they occur.

\clearpage

\noindent\textbf{Acknowledgements.} This work was partially supported by US DoD and DARPA XAI and D3M, NSF awards IIS-1833355, Oculus VR, and the Berkeley Artificial Intelligence Research (BAIR) Lab. DF was supported by a Huawei / Berkeley AI fellowship. Any opinions, findings, and conclusions or recommendations expressed in this material are those of the author(s) and do not necessarily reflect the views of the sponsors, and no official endorsement should be inferred.

\small
\bibliographystyle{ieee}
\bibliography{reference}

\begin{thebibliography}{10}\itemsep=-1pt

\bibitem{anderson2018cvpr}
P.~Anderson, Q.~Wu, D.~Teney, J.~Bruce, M.~Johnson, N.~S{\"u}nderhauf, I.~Reid,
  S.~Gould, and A.~v.~d. Hengel.
\newblock Vision-and-language navigation: Interpreting visually-grounded
  navigation instructions in real environments.
\newblock In {\em Proceedings of the IEEE Conference on Computer Vision and
  Pattern Recognition (CVPR)}, 2018.

\bibitem{Andreas15Instructions}
J.~Andreas and D.~Klein.
\newblock Alignment-based compositional semantics for instruction following.
\newblock In {\em Proceedings of the Conference on Empirical Methods in Natural
  Language Processing (EMNLP)}, 2015.

\bibitem{andreas2016reasoning}
J.~Andreas and D.~Klein.
\newblock Reasoning about pragmatics with neural listeners and speakers.
\newblock In {\em Proceedings of the Conference on Empirical Methods in Natural
  Language Processing (EMNLP)}, 2016.

\bibitem{artzi2013instructions}
Y.~Artzi and L.~Zettlemoyer.
\newblock Weakly supervised learning of semantic parsers for mapping
  instructions to actions.
\newblock {\em Transactions of the Association for Computational Linguistics},
  1(1):49--62, 2013.

\bibitem{bahdanau2014neural}
D.~Bahdanau, K.~Cho, and Y.~Bengio.
\newblock Neural machine translation by jointly learning to align and
  translate.
\newblock In {\em Proceedings of the International Conference on Learning
  Representations (ICLR)}, 2015.

\bibitem{blum1998combining}
A.~Blum and T.~Mitchell.
\newblock Combining labeled and unlabeled data with co-training.
\newblock In {\em Proceedings of the eleventh annual conference on
  Computational learning theory}, pages 92--100. ACM, 1998.

\bibitem{Branavan09PG}
S.~Branavan, H.~Chen, L.~S. Zettlemoyer, and R.~Barzilay.
\newblock Reinforcement learning for mapping instructions to actions.
\newblock In {\em Proceedings of the Annual Meeting of the Association for
  Computational Linguistics (ACL)}, pages 82--90. Association for Computational
  Linguistics, 2009.

\bibitem{Matterport3D}
A.~Chang, A.~Dai, T.~Funkhouser, M.~Halber, M.~Niessner, M.~Savva, S.~Song,
  A.~Zeng, and Y.~Zhang.
\newblock Matterport3d: Learning from rgb-d data in indoor environments.
\newblock {\em International Conference on 3D Vision (3DV)}, 2017.

\bibitem{chen2012grounded}
D.~L. Chen.
\newblock Fast online lexicon learning for grounded language acquisition.
\newblock In {\em Proceedings of the 50th Annual Meeting of the Association for
  Computational Linguistics: Long Papers - Volume 1}, ACL '12, pages 430--439,
  Stroudsburg, PA, USA, 2012. Association for Computational Linguistics.

\bibitem{chen2017query}
K.~Chen, R.~Kovvuri, and R.~Nevatia.
\newblock Query-guided regression network with context policy for phrase
  grounding.
\newblock In {\em Proceedings of the IEEE International Conference on Computer
  Vision (ICCV)}, 2017.

\bibitem{cirik2018using}
V.~Cirik, T.~Berg-Kirkpatrick, and L.-P. Morency.
\newblock Using syntax to ground referring expressions in natural images.
\newblock In {\em 32nd AAAI Conference on Artificial Intelligence (AAAI-18)},
  2018.

\bibitem{cohn2018pragmatically}
R.~Cohn-Gordon, N.~Goodman, and C.~Potts.
\newblock Pragmatically informative image captioning with character-level
  reference.
\newblock In {\em Proceedings of the Conference of the North American Chapter
  of the Association for Computational Linguistics (NAACL)}, 2018.

\bibitem{das2017embodied}
A.~Das, S.~Datta, G.~Gkioxari, S.~Lee, D.~Parikh, and D.~Batra.
\newblock Embodied question answering.
\newblock In {\em Proceedings of the IEEE Conference on Computer Vision and
  Pattern Recognition (CVPR)}, 2018.

\bibitem{frank2012predicting}
M.~C. Frank and N.~D. Goodman.
\newblock Predicting pragmatic reasoning in language games.
\newblock {\em Science}, 336(6084):998--998, 2012.

\bibitem{Frank09PragmaticExperiments}
M.~C. Frank, N.~D. Goodman, P.~Lai, and J.~B. Tenenbaum.
\newblock Informative communication in word production and word learning.
\newblock In {\em {Proceedings of the Annual Conference of the Cognitive
  Science Society}}, 2009.

\bibitem{fried2017unified}
D.~Fried, J.~Andreas, and D.~Klein.
\newblock Unified pragmatic models for generating and following instructions.
\newblock In {\em Proceedings of the Conference of the North American Chapter
  of the Association for Computational Linguistics (NAACL)}, 2018.

\bibitem{goodman2013knowledge}
N.~D. Goodman and A.~Stuhlm{\"u}ller.
\newblock Knowledge and implicature: Modeling language understanding as social
  cognition.
\newblock {\em Topics in cognitive science}, 5(1):173--184, 2013.

\bibitem{Grice75}
H.~P. Grice.
\newblock Logic and conversation.
\newblock In P.~Cole and J.~L. Morgan, editors, {\em Syntax and Semantics: Vol.
  3: Speech Acts}, pages 41--58. Academic Press, San Diego, CA, 1975.

\bibitem{gulcehre2015using}
C.~Gulcehre, O.~Firat, K.~Xu, K.~Cho, L.~Barrault, H.-C. Lin, F.~Bougares,
  H.~Schwenk, and Y.~Bengio.
\newblock On using monolingual corpora in neural machine translation.
\newblock {\em arXiv preprint arXiv:1503.03535}, 2015.

\bibitem{guu2017bridging}
K.~Guu, P.~Pasupat, E.~Z. Liu, and P.~Liang.
\newblock From language to programs: Bridging reinforcement learning and
  maximum marginal likelihood.
\newblock In {\em Proceedings of the Annual Meeting of the Association for
  Computational Linguistics (ACL)}, 2017.

\bibitem{he2016deep}
K.~He, X.~Zhang, S.~Ren, and J.~Sun.
\newblock Deep residual learning for image recognition.
\newblock In {\em Proceedings of the IEEE Conference on Computer Vision and
  Pattern Recognition (CVPR)}, pages 770--778, 2016.

\bibitem{hermann2017grounded}
K.~M. Hermann, F.~Hill, S.~Green, F.~Wang, R.~Faulkner, H.~Soyer,
  D.~Szepesvari, W.~M. Czarnecki, M.~Jaderberg, D.~Teplyashin, M.~Wainwright,
  C.~Apps, D.~Hassabis, and P.~Blunsom.
\newblock Grounded language learning in a simulated 3d world.
\newblock {\em CoRR}, abs/1706.06551, 2017.

\bibitem{hochreiter1997long}
S.~Hochreiter and J.~Schmidhuber.
\newblock Long short-term memory.
\newblock {\em Neural computation}, 9(8):1735--1780, 1997.

\bibitem{hu17cvpr}
R.~Hu, M.~Rohrbach, J.~Andreas, T.~Darrell, and K.~Saenko.
\newblock Modeling relationships in referential expressions with compositional
  modular networks.
\newblock In {\em Proceedings of the IEEE Conference on Computer Vision and
  Pattern Recognition (CVPR)}, 2017.

\bibitem{hu16eccv}
R.~Hu, M.~Rohrbach, and T.~Darrell.
\newblock Segmentation from natural language expressions.
\newblock In {\em Proceedings of the European Conference on Computer Vision
  (ECCV)}, 2016.

\bibitem{hu16cvpr}
R.~Hu, H.~Xu, M.~Rohrbach, J.~Feng, K.~Saenko, and T.~Darrell.
\newblock Natural language object retrieval.
\newblock In {\em Proceedings of the IEEE Conference on Computer Vision and
  Pattern Recognition (CVPR)}, 2016.

\bibitem{kovcisky2016semisupervised}
T.~Ko\v{c}isk\'{y}, G.~Melis, E.~Grefenstette, C.~Dyer, W.~Ling, P.~Blunsom,
  and K.~M. Hermann.
\newblock Semantic parsing with semi-supervised sequential autoencoders.
\newblock In {\em Proceedings of the 2016 Conference on Empirical Methods in
  Natural Language Processing}, pages 1078--1087, Austin, Texas, November 2016.
  Association for Computational Linguistics.

\bibitem{liu2017recurrent}
C.~Liu, Z.~Lin, X.~Shen, J.~Yang, X.~Lu, and A.~Yuille.
\newblock Recurrent multimodal interaction for referring image segmentation.
\newblock In {\em Proceedings of the IEEE International Conference on Computer
  Vision (ICCV)}, 2017.

\bibitem{long2016simpler}
R.~Long, P.~Pasupat, and P.~Liang.
\newblock Simpler context-dependent logical forms via model projections.
\newblock In {\em Proceedings of the Annual Meeting of the Association for
  Computational Linguistics (ACL)}, 2016.

\bibitem{luo17cvpr}
R.~Luo and G.~Shakhnarovich.
\newblock Comprehension-guided referring expressions.
\newblock In {\em Proceedings of the IEEE Conference on Computer Vision and
  Pattern Recognition (CVPR)}, 2017.

\bibitem{mao2016generation}
J.~Mao, H.~Jonathan, A.~Toshev, O.~Camburu, A.~Yuille, and K.~Murphy.
\newblock Generation and comprehension of unambiguous object descriptions.
\newblock In {\em Proceedings of the IEEE Conference on Computer Vision and
  Pattern Recognition (CVPR)}, 2016.

\bibitem{mcclosky2006effective}
D.~McClosky, E.~Charniak, and M.~Johnson.
\newblock Effective self-training for parsing.
\newblock In {\em Proceedings of the main conference on human language
  technology conference of the North American Chapter of the Association of
  Computational Linguistics}, pages 152--159. Association for Computational
  Linguistics, 2006.

\bibitem{Mei16Instructions}
H.~Mei, M.~Bansal, and M.~Walter.
\newblock Listen, attend, and walk: Neural mapping of navigational instructions
  to action sequences.
\newblock In {\em Proceedings of the Conference on Artificial Intelligence
  (AAAI)}, 2016.

\bibitem{misra2017instructions}
D.~Misra, J.~Langford, and Y.~Artzi.
\newblock Mapping instructions and visual observations to actions with
  reinforcement learning.
\newblock In {\em Proceedings of the Conference on Empirical Methods in Natural
  Language Processing (EMNLP)}, 2017.

\bibitem{monroe2017colors}
W.~Monroe, R.~Hawkins, N.~Goodman, and C.~Potts.
\newblock Colors in context: A pragmatic neural model for grounded language
  understanding.
\newblock {\em Transactions of the Association for Computational Linguistics},
  5:325--338, 2017.

\bibitem{nagaraja16eccv}
V.~K. Nagaraja, V.~I. Morariu, and L.~S. Davis.
\newblock Modeling context between objects for referring expression
  understanding.
\newblock In {\em Proceedings of the European Conference on Computer Vision
  (ECCV)}, pages 792--807. Springer, 2016.

\bibitem{pathak2018zero}
D.~Pathak, P.~Mahmoudieh, G.~Luo, P.~Agrawal, D.~Chen, Y.~Shentu, E.~Shelhamer,
  J.~Malik, A.~A. Efros, and T.~Darrell.
\newblock Zero-shot visual imitation.
\newblock {\em arXiv preprint arXiv:1804.08606}, 2018.

\bibitem{pennington2014glove}
J.~Pennington, R.~Socher, and C.~Manning.
\newblock Glove: Global vectors for word representation.
\newblock In {\em Proceedings of the 2014 conference on empirical methods in
  natural language processing (EMNLP)}, pages 1532--1543, 2014.

\bibitem{plummer15iccv}
B.~Plummer, L.~Wang, C.~Cervantes, J.~Caicedo, J.~Hockenmaier, and S.~Lazebnik.
\newblock Flickr30k entities: Collecting region-to-phrase correspondences for
  richer image-to-sentence models.
\newblock In {\em Proceedings of the IEEE International Conference on Computer
  Vision (ICCV)}, 2015.

\bibitem{radosavovic2017data}
I.~Radosavovic, P.~Doll{\'a}r, R.~Girshick, G.~Gkioxari, and K.~He.
\newblock Data distillation: Towards omni-supervised learning.
\newblock {\em arXiv preprint arXiv:1712.04440}, 2017.

\bibitem{rohrbach16eccv}
A.~Rohrbach, M.~Rohrbach, R.~Hu, T.~Darrell, and B.~Schiele.
\newblock Grounding of textual phrases in images by reconstruction.
\newblock In {\em Proceedings of the European Conference on Computer Vision
  (ECCV)}, 2016.

\bibitem{russakovsky2015imagenet}
O.~Russakovsky, J.~Deng, H.~Su, J.~Krause, S.~Satheesh, S.~Ma, Z.~Huang,
  A.~Karpathy, A.~Khosla, M.~Bernstein, et~al.
\newblock Imagenet large scale visual recognition challenge.
\newblock {\em International Journal of Computer Vision}, 115(3):211--252,
  2015.

\bibitem{scudder1965probability}
H.~Scudder.
\newblock Probability of error of some adaptive pattern-recognition machines.
\newblock {\em IEEE Transactions on Information Theory}, 11(3):363--371, 1965.

\bibitem{sennrich2016mt}
R.~Sennrich, B.~Haddow, and A.~Birch.
\newblock Improving neural machine translation models with monolingual data.
\newblock In {\em Proceedings of the Annual Meeting of the Association for
  Computational Linguistics (ACL)}, pages 86--96, 2016.

\bibitem{silver2017mastering}
D.~Silver, T.~Hubert, J.~Schrittwieser, I.~Antonoglou, M.~Lai, A.~Guez,
  M.~Lanctot, L.~Sifre, D.~Kumaran, T.~Graepel, et~al.
\newblock Mastering chess and shogi by self-play with a general reinforcement
  learning algorithm.
\newblock {\em arXiv preprint arXiv:1712.01815}, 2017.

\bibitem{smith2013learning}
N.~J. Smith, N.~Goodman, and M.~Frank.
\newblock Learning and using language via recursive pragmatic reasoning about
  other agents.
\newblock In {\em Advances in neural information processing systems}, pages
  3039--3047, 2013.

\bibitem{sukhbaatar2017intrinsic}
S.~Sukhbaatar, Z.~Lin, I.~Kostrikov, G.~Synnaeve, A.~Szlam, and R.~Fergus.
\newblock Intrinsic motivation and automatic curricula via asymmetric
  self-play.
\newblock {\em arXiv preprint arXiv:1703.05407}, 2017.

\bibitem{sutskever2014sequence}
I.~Sutskever, O.~Vinyals, and Q.~V. Le.
\newblock Sequence to sequence learning with neural networks.
\newblock In {\em Advances in neural information processing systems}, pages
  3104--3112, 2014.

\bibitem{sutton1998reinforcement}
R.~S. Sutton and A.~G. Barto.
\newblock {\em Reinforcement learning: An introduction}, volume~1.
\newblock MIT press Cambridge, 1998.

\bibitem{sutton2000policy}
R.~S. Sutton, D.~A. McAllester, S.~P. Singh, and Y.~Mansour.
\newblock Policy gradient methods for reinforcement learning with function
  approximation.
\newblock In {\em Advances in neural information processing systems}, pages
  1057--1063, 2000.

\bibitem{tellex2011understanding}
S.~Tellex, T.~Kollar, S.~Dickerson, M.~R. Walter, A.~G. Banerjee, S.~J. Teller,
  and N.~Roy.
\newblock Understanding natural language commands for robotic navigation and
  mobile manipulation.
\newblock In {\em AAAI}, volume~1, page~2, 2011.

\bibitem{vasudevan2018referring}
A.~B. Vasudevan, D.~Dai, and L.~V. Gool.
\newblock Object referring in visual scene with spoken language.
\newblock In {\em Proc.~IEEE Winter Conf.~on Applications of Computer Vision
  (WACV)}, 2018.

\bibitem{vedantam2017context}
R.~Vedantam, S.~Bengio, K.~Murphy, D.~Parikh, and G.~Chechik.
\newblock Context-aware captions from context-agnostic supervision.
\newblock In {\em Proceedings of the IEEE Conference on Computer Vision and
  Pattern Recognition (CVPR)}, volume~3, 2017.

\bibitem{wang16eccv}
M.~Wang, M.~Azab, N.~Kojima, R.~Mihalcea, and J.~Deng.
\newblock Structured matching for phrase localization.
\newblock In {\em Proceedings of the European Conference on Computer Vision
  (ECCV)}, pages 696--711. Springer, 2016.

\bibitem{wang2018look}
X.~Wang, W.~Xiong, H.~Wang, and W.~Y. Wang.
\newblock Look before you leap: Bridging model-free and model-based
  reinforcement learning for planned-ahead vision-and-language navigation.
\newblock {\em arXiv:1803.07729}, 2018.

\bibitem{weber2017imagination}
T.~Weber, S.~Racani{\`e}re, D.~P. Reichert, L.~Buesing, A.~Guez, D.~J. Rezende,
  A.~P. Badia, O.~Vinyals, N.~Heess, Y.~Li, et~al.
\newblock Imagination-augmented agents for deep reinforcement learning.
\newblock {\em arXiv preprint arXiv:1707.06203}, 2017.

\bibitem{yu2018mattnet}
L.~Yu, Z.~Lin, X.~Shen, J.~Yang, X.~Lu, M.~Bansal, and T.~L. Berg.
\newblock Mattnet: Modular attention network for referring expression
  comprehension.
\newblock In {\em Proceedings of the IEEE Conference on Computer Vision and
  Pattern Recognition (CVPR)}, 2018.

\bibitem{yu2017joint}
L.~Yu, H.~Tan, M.~Bansal, and T.~L. Berg.
\newblock A joint speaker-listener-reinforcer model for referring expressions.
\newblock In {\em Proceedings of the IEEE Conference on Computer Vision and
  Pattern Recognition (CVPR)}, 2017.

\end{thebibliography}

\clearpage
\vspace{4em}
\begin{figure}[ht!]
  \centering
  \Large\textbf{Supplementary Material}
\end{figure}
\vspace{-1em}

\appendix
\setcounter{table}{0}
\renewcommand{\thetable}{\Alph{section}.\arabic{table}}
\setcounter{figure}{0}
\renewcommand{\thefigure}{\Alph{section}.\arabic{figure}}
\setcounter{equation}{0}
\renewcommand{\theequation}{\Alph{section}.\arabic{equation}}
\setcounter{algorithm}{0}
\renewcommand{\thealgorithm}{\Alph{section}.\arabic{algorithm}}

\normalsize
\section{Overview}

In this document we describe our algorithm for candidate route generation, and provide analysis on parameters for speaker-driven route selection (pragmatic inference) and other details. We also provide additional qualitative examples showing how pragmatic inference helps instruction following.
Finally, we describe our submission to the Vision and Language Navigation Challenge.

\section{State-Factored Search for Candidate Route Generation}
\label{sec:supp_state_factored_search}


Algorithm~\ref{alg:supp_state_factored} gives pseudocode for the state-factored search algorithm that we use to generate candidate routes for pragmatic inference (outlined in Sec.~3.2 of the main paper).

\algblockdefx{Class}{EndClass}%
    [1]{\textbf{class} \textsc{#1} }%
    [0]{\textbf{end class} }

\begin{algorithm}  
  \caption{State-factored search
    \label{alg:supp_state_factored}}
  \begin{algorithmic}[5]  
    \Class{State}
        \State $location$ \Comment{the agent's physical position in the environment}
        \State $completed$ \Comment{whether the route has been completed (the \textsc{Stop} action has been taken)}
    \EndClass \\
    \Class{Route}
        \State $states$ \Comment{list of \textsc{State}s in the route}
        \State $score$ \Comment{route probability under the follower model, $P_F$}
        \State $expanded$ \Comment{whether this route has been expanded in search}
    \EndClass \\
    \Function{StateFactoredSearch}{$start\_state: \textsc{State}$, $K: int$}  
      \LineComment{a mapping from \textsc{State}s to the best \textsc{Route} ending in that state found so far (whether expanded or unexpanded)}
      \State $partial = \{\}$
      \
      \LineComment{a similar mapping, but containing routes that have been completed}
      \State $completed = \{\}$
      \State $start\_route = \textsc{Route}([start\_state], 1.0, False)$
      \State $partial[start\_state] = start\_route$
      \State $candidates = [start\_route]$
      \While{$|completed| < K \textbf{ and } |candidates| > 0$}
        \LineComment{choose the highest-scoring unexpanded route to expand (route may be complete)}
        \State $route = argmax_{r \in candidates } r.score$
        \State $route.expanded = True$
        \LineComment{\textsc{Successor} generates \textsc{Route}s by taking all possible actions, each of which extends this route by one state, with the resulting total model $score$, and $expanded$ set to False}
        \For{$route' \textbf{ in } \textsc{Successors}(route)$} 
          \State $state' = route'.states.last$
          \State $cache = completed \textbf{ if } route'.completed \textbf{ else } partial$
          \If{$state' \textbf{ not in } cache.keys \textbf{ or } cache[state'].score < route'.score$}
            \State $cache[state'] = route'$
          \EndIf
        \EndFor
        \State $candidates = [route \textbf{ in } partial.values \textbf{ if not } route.expanded ]$
      \EndWhile
      \State \Return{$completed$}
    \EndFunction  
  \end{algorithmic}  
\end{algorithm}

\section{Analysis of Inference Parameters and Implementation Details}
\label{sec:parameter_analysis}

\begin{figure}[t]
\centering
\includegraphics[width=0.75\linewidth]{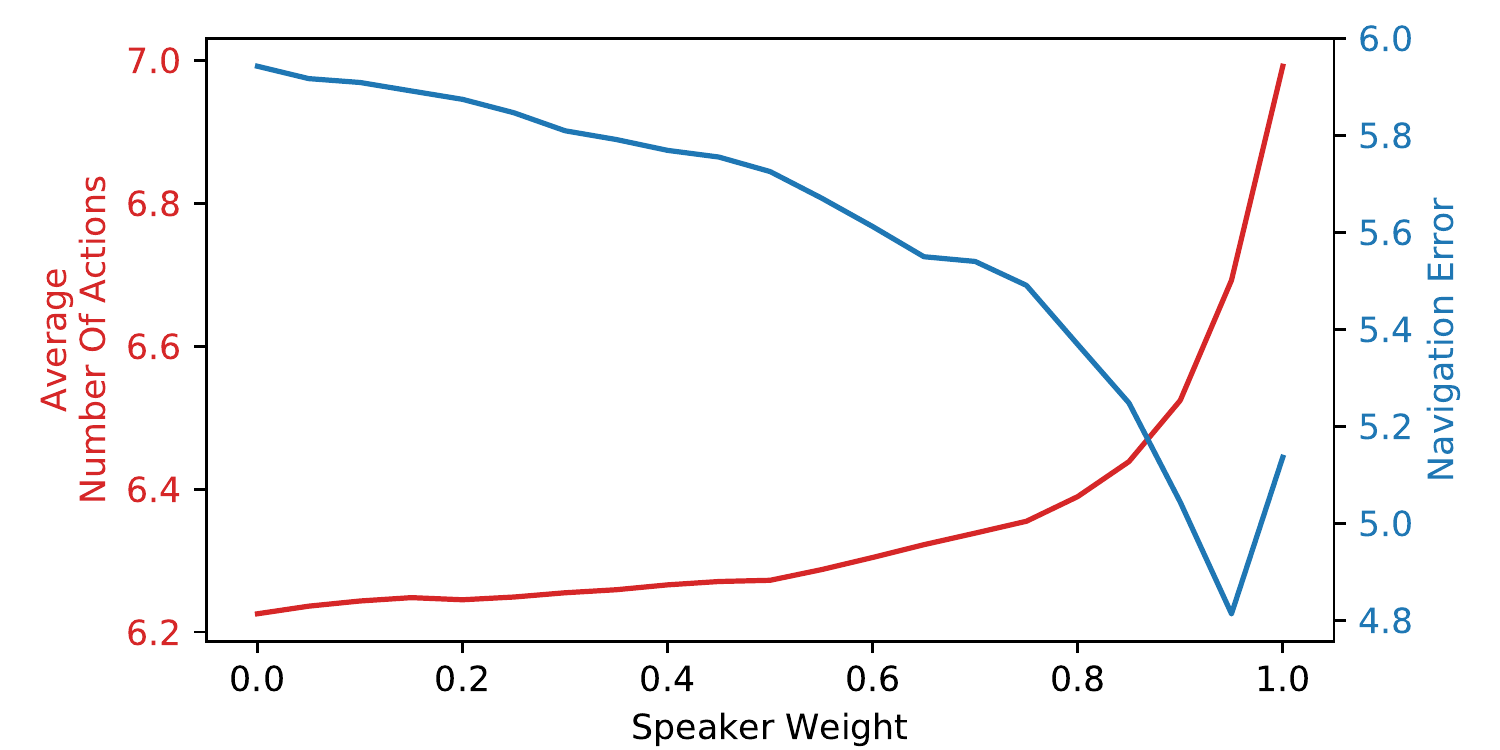}
\caption{The average number of actions and navigation error with different speaker weights $\lambda$ in pragmatic inference (Sec.~3.2 of the main paper), evaluated on the val unseen split. Larger $\lambda$ results in more number of actions on average, while $\lambda=0.95$ gives the lowest navigation error.}
\label{fig:supp_speaker_weight}
\end{figure}

We further study how varying the speaker weight $\lambda$ in pragmatic inference (Sec.~3.2 of the main paper) influences the final navigation performance. In Table~\ref{tab:supp_ablations}, we compare our full model (with speaker weight $\lambda=0.95$) in Row 1 against using only the follower model to score routes ($\lambda=0$) in Row 2, a baseline that still includes search but does not include speaker scores. The large gap in success rate between $\lambda=0.95$ and $\lambda=0$ shows that including speaker scores is crucial to the performance, confirming the importance of pragmatic inference. Figure~\ref{fig:supp_speaker_weight} shows the average number of actions and the navigation error on val unseen with different speaker weights $\lambda$, where $\lambda=0.95$ gives the lowest navigation error.

\begin{figure}[t]
\centering
\includegraphics[width=0.7\linewidth]{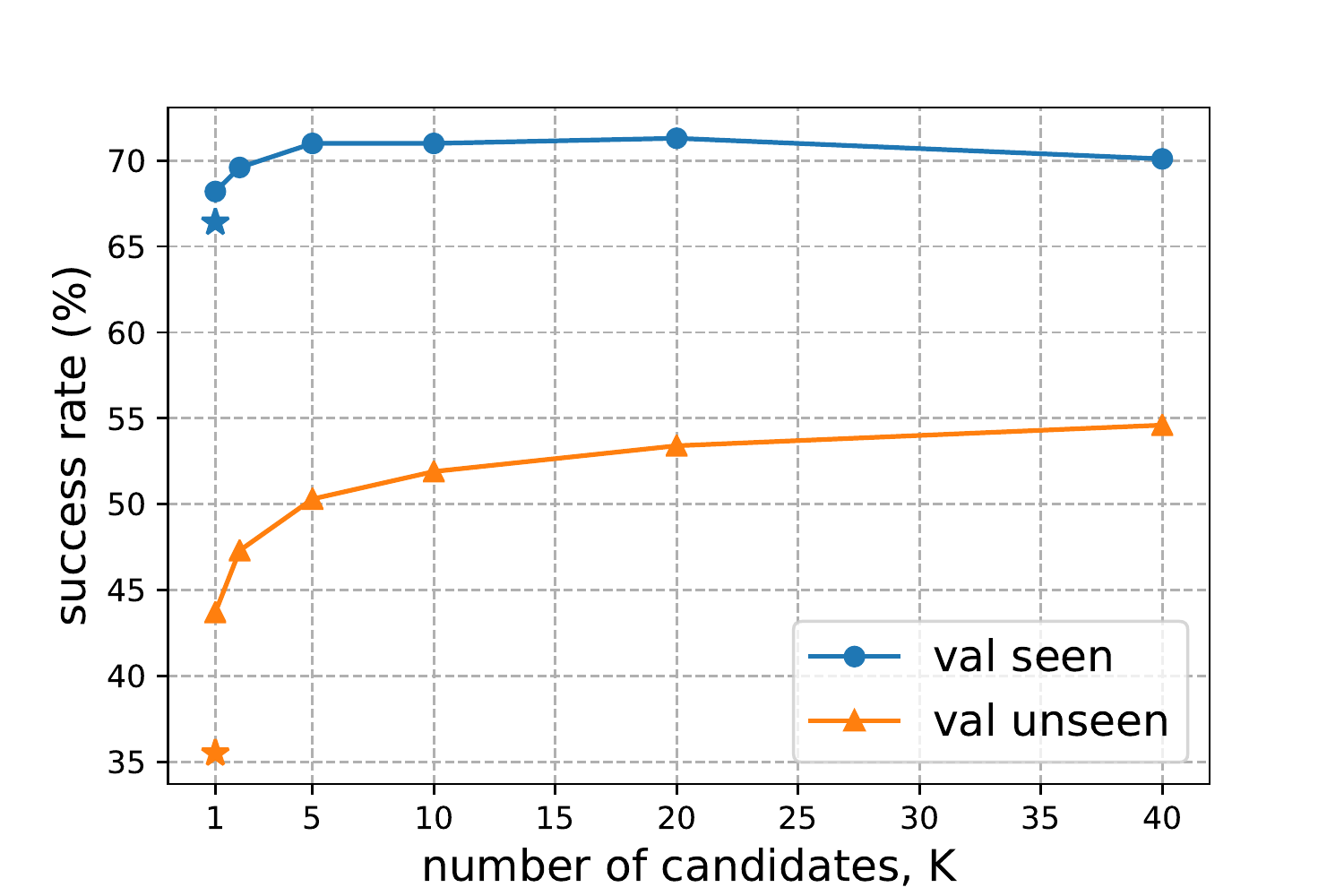}
\caption{The success rate of our model on the val seen and val unseen splits, using different numbers $K$ of route candidates (generated by state-factored search) for pragmatic inference. Stars show the performance of greedy inference (without search, and hence without pragmatics). While performance increases with number of candidates up through 40 on val unseen, the success rate tends to saturate. We note improvements both from the state-factored search procedure (comparing the stars to the circle and triangle points at $K=1$) as well as from having more candidates to choose from in pragmatic inference (comparing larger values of $K$ to smaller). \vspace{-1em}}
\label{fig:supp_beam_size}
\end{figure}

In addition, we study how the number of candidate routes, $K$, used in pragmatic inference impacts the final navigation success rate. Figure~\ref{fig:supp_beam_size} shows the success rate of our model on R2R val seen and val unseen splits using different numbers of candidate routes $K$ for state-factored search (Sec.~\ref{sec:supp_state_factored_search}). The results show that having more routes to choose between leads to better generalization to the unseen new environments (val unseen), but the gain from increasing the number of candidates tends to saturate quickly. In our final model in Table~2 in the main paper and Table~\ref{tab:supp_ablations}, we use $K=40$. However, we emphasize that \textbf{even with only five route candidates, our model still achieves 50.3\% success rate on val unseen}, which improves substantially on both the $35.5\%$ success rate from greedy decoding (i.e.\ the gold star at $K=1$ in Figure~\ref{fig:supp_beam_size}), as well as the 43.5\% success rate given by state-factored search with no pragmatic inference (i.e.\ the gold triangle at $K=1$ in Figure~\ref{fig:supp_beam_size}).

\begin{table}[t]
\small
\begin{center}
\begin{tabular}{rlcccccc}
\toprule
  &  & \multicolumn{3}{c}{Validation-Seen} & \multicolumn{3}{c}{Validation-Unseen} \\
  \cmidrule(l){3-5}\cmidrule(l){6-8}
  \# &  & NE $\downarrow$ & SR $\uparrow$ & OSR $\uparrow$ & NE $\downarrow$ & SR $\uparrow$ & OSR $\uparrow$ \\
\cmidrule(l){1-1}\cmidrule(l){2-2}\cmidrule(l){3-3}\cmidrule(l){4-4}\cmidrule(l){5-5}\cmidrule(l){6-6}\cmidrule(l){7-7}\cmidrule(l){8-8}
  1 & our full model ($\lambda=0.95$) & \textbf{3.08} & 70.1 & \textbf{78.3} & \textbf{4.83} & \textbf{54.6} & 65.2 \\
  \cmidrule(r){1-8}
  2 & w/o speaker scoring ($\lambda = 0$) & 3.17 & 68.4 & 74.8 & 5.94 & 43.7 & 53.1 \\
  3 & w/o state-factoring in search (Sec.\ \ref{sec:supp_state_factored_search}) & 3.14 & \textbf{70.6} & 77.4 & 5.27 & 50.7 & 60.7 \\
  4 & w/o GloVe embedding \cite{pennington2014glove} & 3.08 & 69.6 & 77.4 & 4.84 & 53.2 & \textbf{66.7} \\
\bottomrule
\end{tabular}
\end{center}
\caption{Effects of speaker scoring and implementation details in our model. NE is navigation error (in meters); lower is better. SR and OSR are success rate and oracle success rate (\%); higher is better. Comparison between the 1st and the 2nd row shows that incorporating speaker scoring is crucial to the performance, matching our intuition in Sec.~3.2 of the main paper of the importance of pragmatic inference. Comparison between the 1st and the 3rd row indicates that state-factored search (Sec.~\ref{sec:supp_state_factored_search}) produces better results than (standard) beam search on val-unseen. Difference between the 1st and the 4th row shows that using GloVe embeddings \cite{pennington2014glove} gives slightly higher success rate.}
\label{tab:supp_ablations}
\end{table}

We also analyze some implementation details in our model. Comparing Row 1 v.s. Row 3 in Table~\ref{tab:supp_ablations} shows that using state-factored search to produce candidates for pragmatic inference (Sec.\ \ref{sec:supp_state_factored_search}) produces better results on val unseen than using (standard) beam search.
Comparing Row 1 v.s. Row 4 in Table~\ref{tab:supp_ablations} indicates that using GloVe \cite{pennington2014glove} to initialize word embedding sightly improves success rate.

\section{Qualitative Examples}

\begin{figure}[t]
\vspace{-2em}
\begin{center}
\textbf{Instruction}:\\
\textit{Walk down and turn right. Walk a bit, and turn right towards the door. Enter inside, and stop in front of a zebra striped rug.} \\ ~ \\
\small{\textit{rear}: -180 degree~~~~~~~~~~~~\textit{left}: -90 degree~~~~~~~~~~~~~~\textit{front}: 0 degree~~~~~~~~~~\textit{right}: +90 degree~~~~~~~~~\textit{rear}: +180 degree} \\
\includegraphics[width=\linewidth,trim={5cm 0 4.3cm 0},clip]{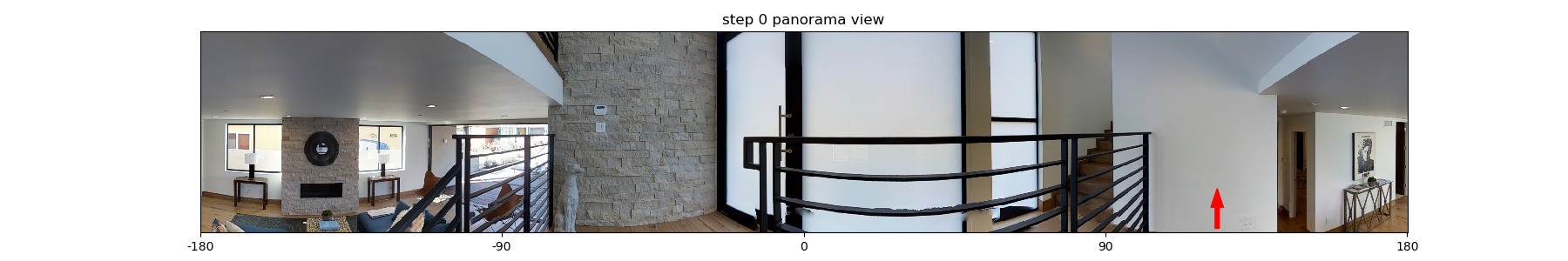}
\includegraphics[width=\linewidth,trim={5cm 0 4.3cm 0},clip]{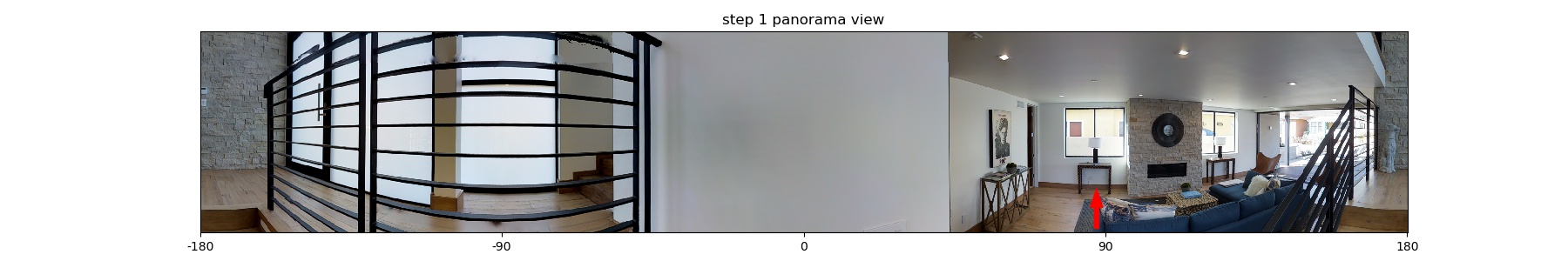}
\includegraphics[width=\linewidth,trim={5cm 0 4.3cm 0},clip]{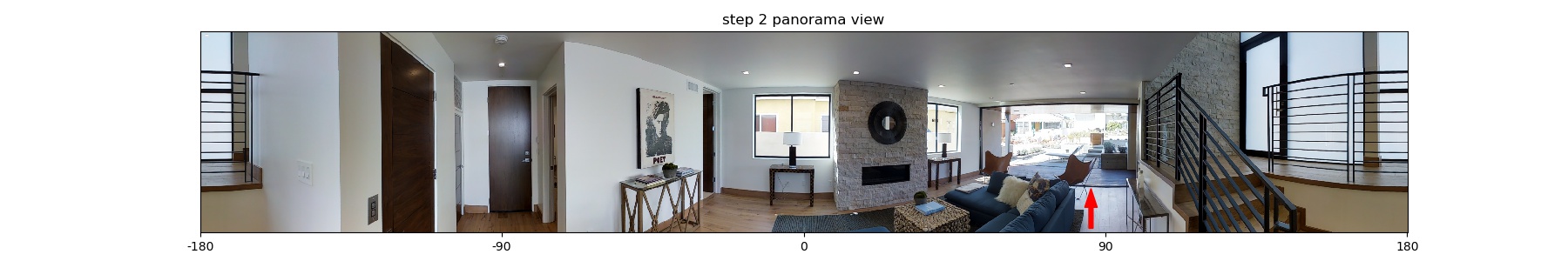}
\includegraphics[width=\linewidth,trim={5cm 0 4.3cm 0},clip]{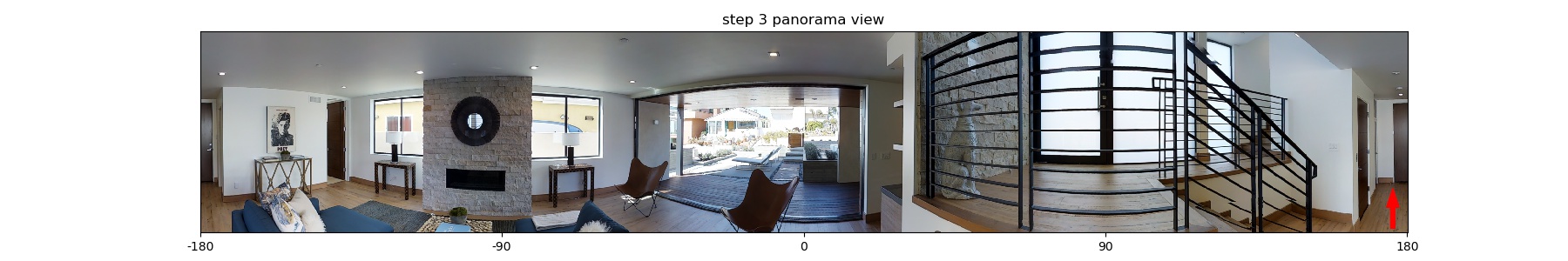}
\includegraphics[width=\linewidth,trim={5cm 0 4.3cm 0},clip]{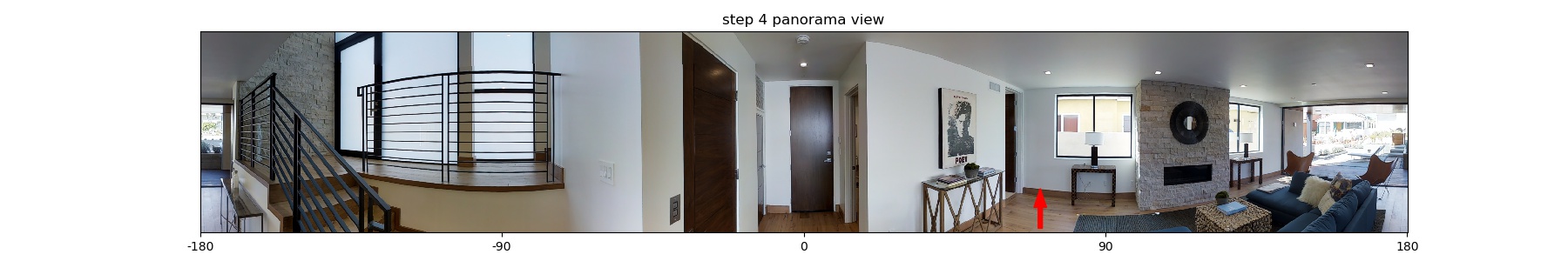}
\includegraphics[width=\linewidth,trim={5cm 0 4.3cm 0},clip]{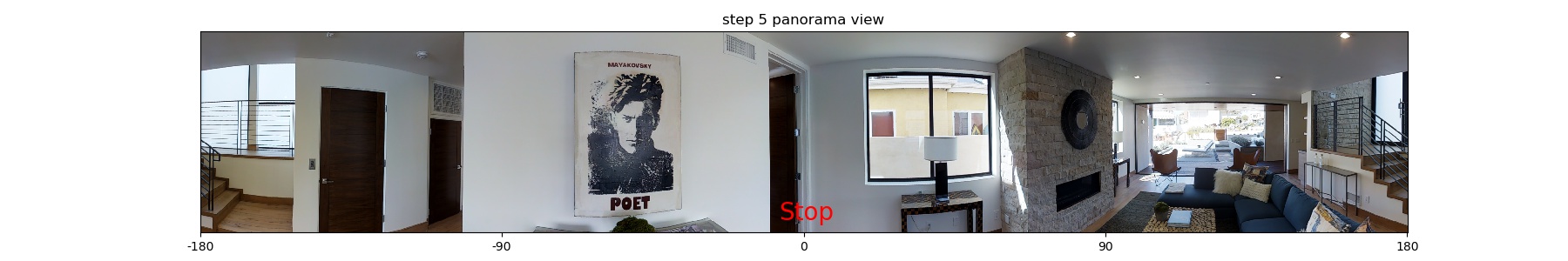}
\\ \small{Navigation steps of the panorama agent. The \textcolor{red}{red} arrow shows the direction chosen by the agent to go next.} \\
\end{center}
\caption{Follower \textbf{without pragmatic inference} on val seen. The instruction involves an ambiguous ``walk a bit'' command. Without pragmatic reasoning by the speaker, the follower failed to predict how much to move forward, stopping at a wrong location without entering the door.}
\label{fig:supp_base_seen}
\vspace{-2em}
\end{figure}

\begin{figure}[t]
\vspace{-2em}
\begin{center}
\textbf{Instruction}:\\
\textit{Walk down and turn right. Walk a bit, and turn right towards the door. Enter inside, and stop in front of a zebra striped rug.} \\ ~ \\
\small{\textit{rear}: -180 degree~~~~~~~~~~~~\textit{left}: -90 degree~~~~~~~~~~~~~~\textit{front}: 0 degree~~~~~~~~~~\textit{right}: +90 degree~~~~~~~~~\textit{rear}: +180 degree} \\
\includegraphics[width=0.8\linewidth,trim={5cm 0.2cm 4.3cm 0.2cm},clip]{figures/58_4932_1/0.jpg}
\includegraphics[width=0.8\linewidth,trim={5cm 0.2cm 4.3cm 0.2cm},clip]{figures/58_4932_1/1.jpg}
\includegraphics[width=0.8\linewidth,trim={5cm 0.2cm 4.3cm 0.2cm},clip]{figures/58_4932_1/2.jpg}
\includegraphics[width=0.8\linewidth,trim={5cm 0.2cm 4.3cm 0.2cm},clip]{figures/58_4932_1/3.jpg}
\includegraphics[width=0.8\linewidth,trim={5cm 0.2cm 4.3cm 0.2cm},clip]{figures/58_4932_1/4.jpg}
\includegraphics[width=0.8\linewidth,trim={5cm 0.2cm 4.3cm 0.2cm},clip]{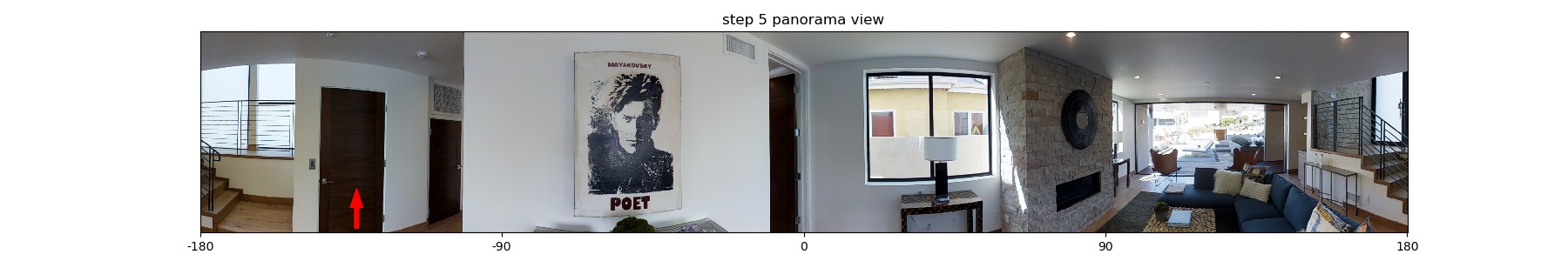}
\includegraphics[width=0.8\linewidth,trim={5cm 0.2cm 4.3cm 0.2cm},clip]{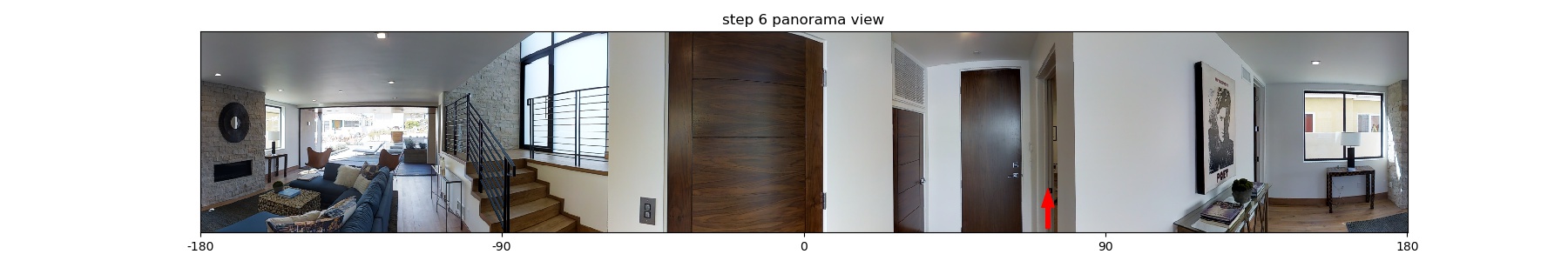}
\includegraphics[width=0.8\linewidth,trim={5cm 0.2cm 4.3cm 0.2cm},clip]{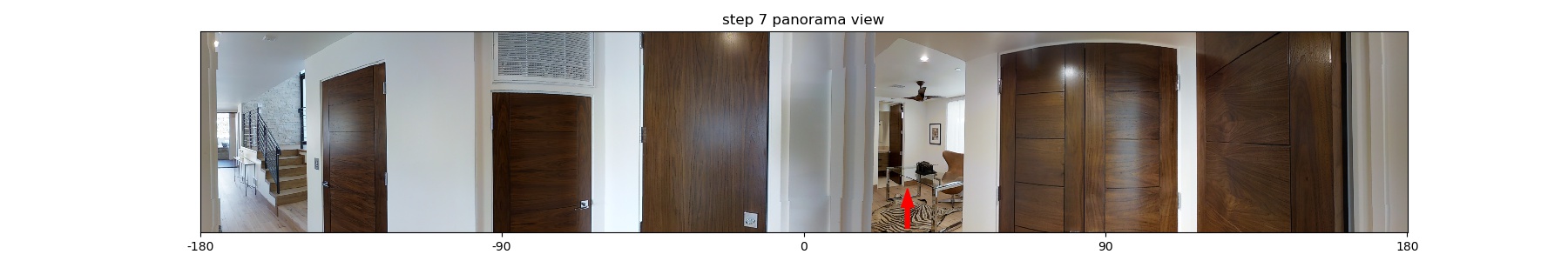}
\includegraphics[width=0.8\linewidth,trim={5cm 0.2cm 4.3cm 0.2cm},clip]{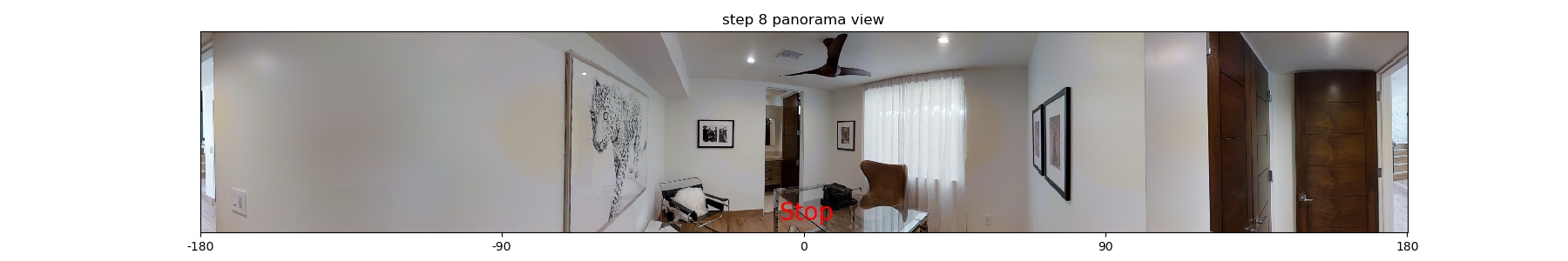}
\\ \small{Navigation steps of the panorama agent. The \textcolor{red}{red} arrow shows the direction chosen by the agent to go next.} \\
\end{center}
\vspace{-1em}
\caption{Follower \textbf{with pragmatic inference} on val seen. With the help of the speaker, the follower could disambiguate ``walk a bit'' to move the right amount to the correct location. It then turned right and walked into the door to stop by the ``zebra stripped rug''.}
\label{fig:supp_rational_seen}
\vspace{-2em}
\end{figure}

\begin{figure}[t]
\vspace{-2em}
\begin{center}
\textbf{Instruction}:\\
\textit{Walk past hall table. Walk into bedroom. Make left at table clock. Wait at bathroom door threshold.} \\ ~ \\
\small{\textit{rear}: -180 degree~~~~~~~~~~~~\textit{left}: -90 degree~~~~~~~~~~~~~~\textit{front}: 0 degree~~~~~~~~~~\textit{right}: +90 degree~~~~~~~~~\textit{rear}: +180 degree} \\
\includegraphics[width=\linewidth,trim={5cm 0 4.3cm 0},clip]{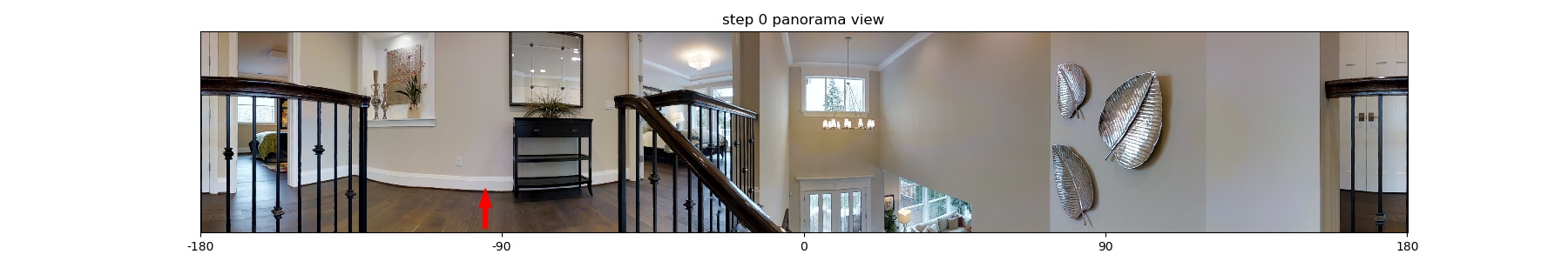}
\includegraphics[width=\linewidth,trim={5cm 0 4.3cm 0},clip]{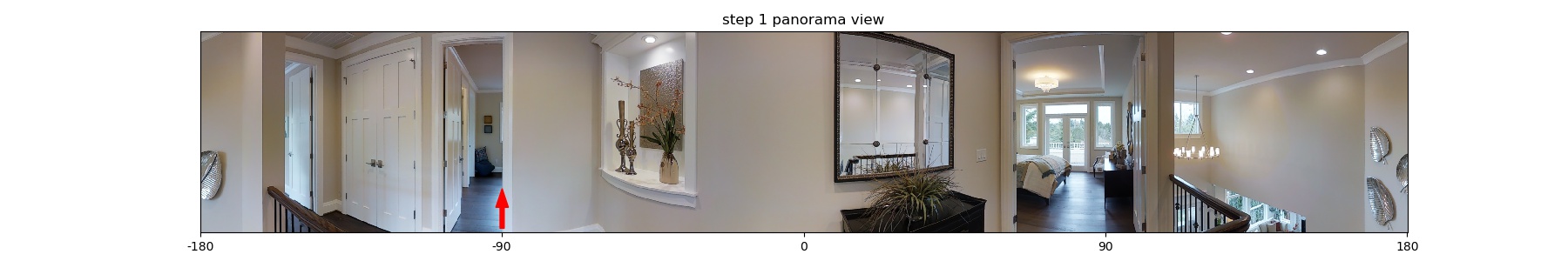}
\includegraphics[width=\linewidth,trim={5cm 0 4.3cm 0},clip]{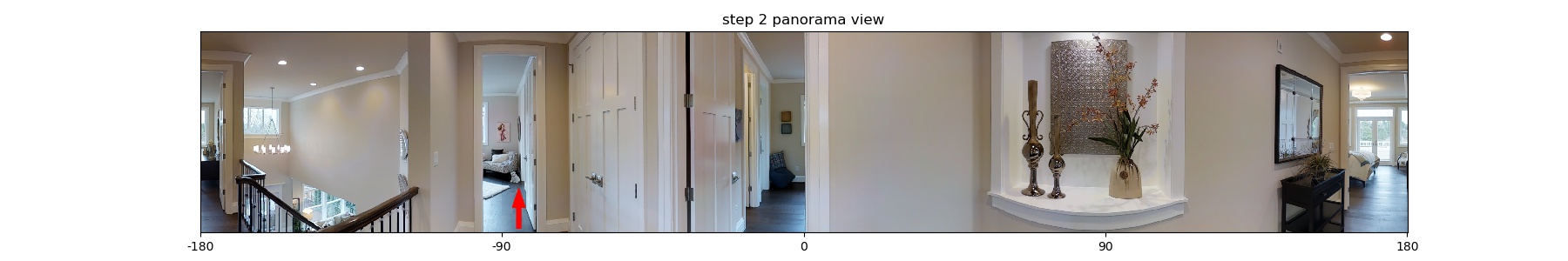}
\includegraphics[width=\linewidth,trim={5cm 0 4.3cm 0},clip]{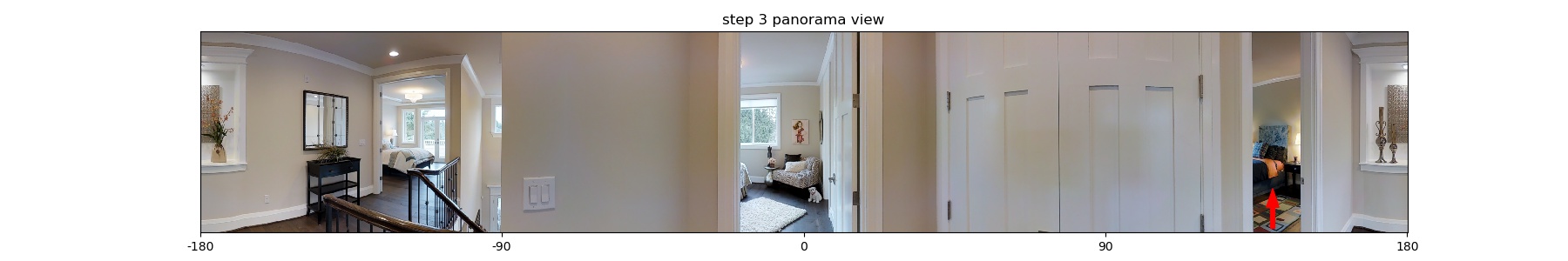}
\includegraphics[width=\linewidth,trim={5cm 0 4.3cm 0},clip]{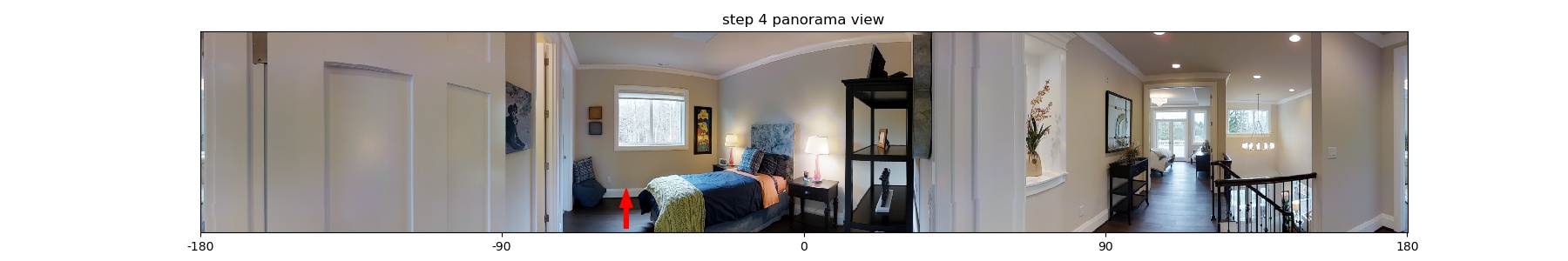}
\includegraphics[width=\linewidth,trim={5cm 0 4.3cm 0},clip]{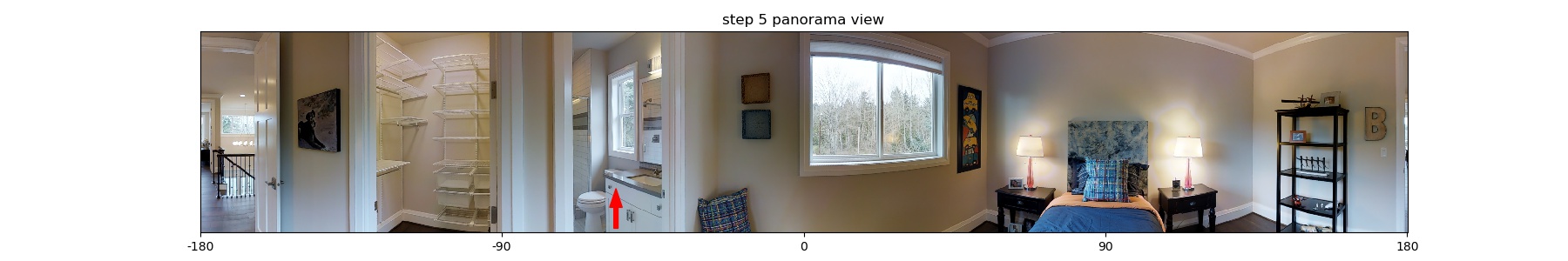}
\includegraphics[width=\linewidth,trim={5cm 0 4.3cm 0},clip]{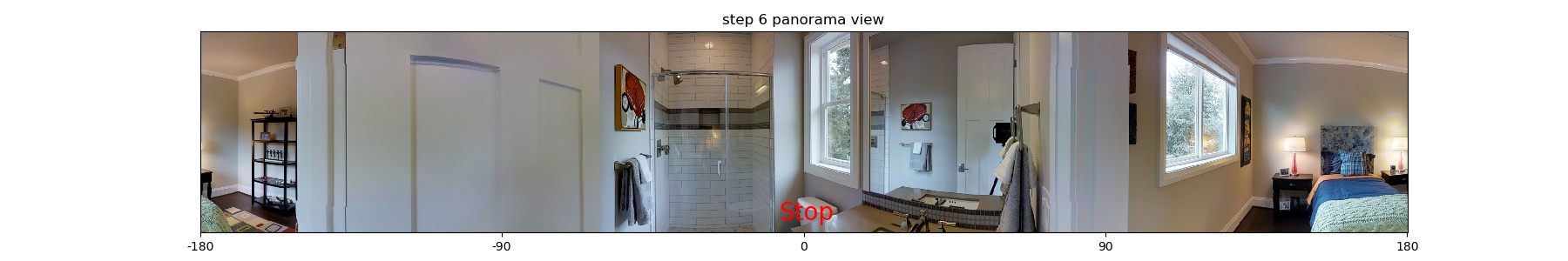}
\\ \small{Navigation steps of the panorama agent. The \textcolor{red}{red} arrow shows the direction chosen by the agent to go next.} \\
\end{center}
\vspace{-1em}
\caption{Follower \textbf{without pragmatic inference} on val unseen. The command ``walk into bedroom'' is ambiguous since there are two bedrooms (one on the left and one on the right). The follower could not decide which bedroom to enter, but went into the wrong room with no ``table clock''.}
\label{fig:supp_base_unseen1}
\vspace{-2em}
\end{figure}

\begin{figure}[t]
\vspace{-2em}
\begin{center}
\textbf{Instruction}:\\
\textit{Walk past hall table. Walk into bedroom. Make left at table clock. Wait at bathroom door threshold.} \\ ~ \\
\small{\textit{rear}: -180 degree~~~~~~~~~~~~\textit{left}: -90 degree~~~~~~~~~~~~~~\textit{front}: 0 degree~~~~~~~~~~\textit{right}: +90 degree~~~~~~~~~\textit{rear}: +180 degree} \\
\includegraphics[width=\linewidth,trim={5cm 0 4.3cm 0},clip]{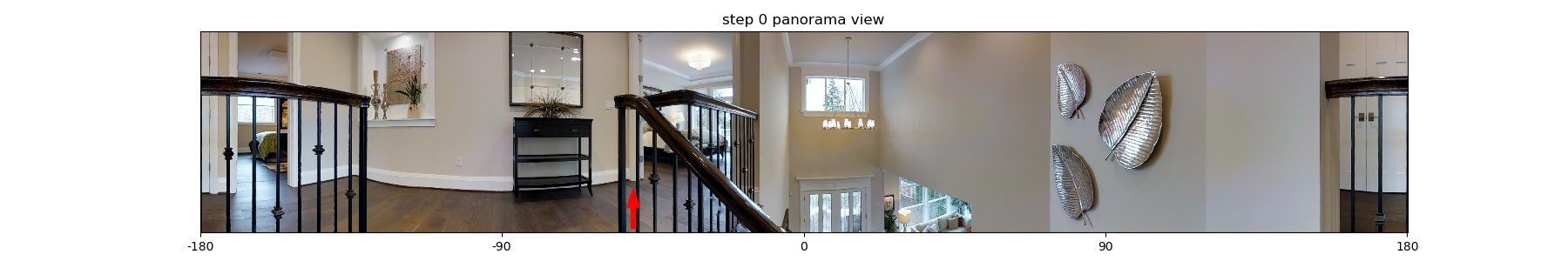}
\includegraphics[width=\linewidth,trim={5cm 0 4.3cm 0},clip]{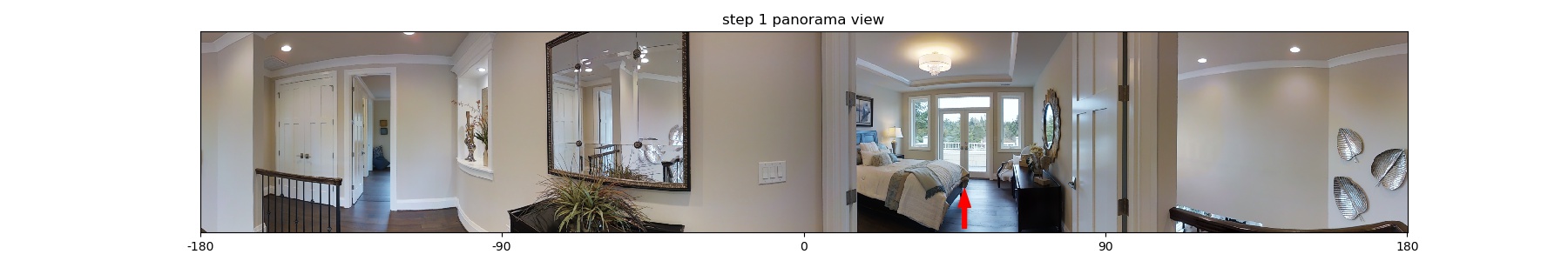}
\includegraphics[width=\linewidth,trim={5cm 0 4.3cm 0},clip]{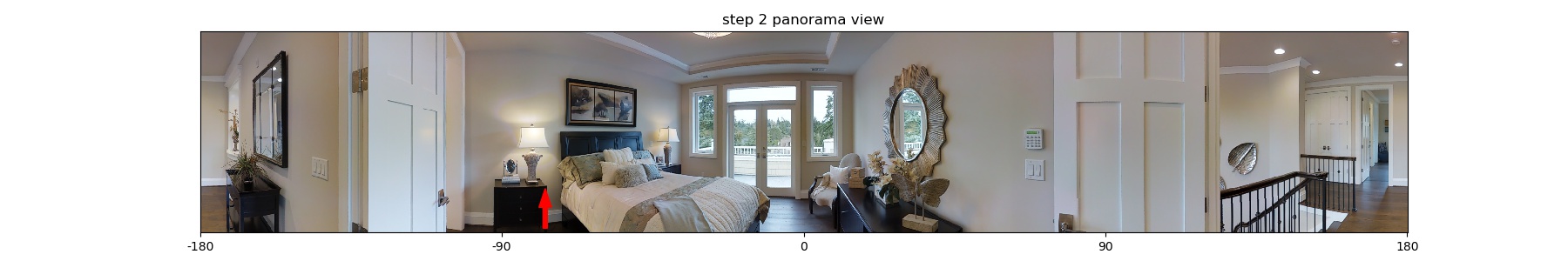}
\includegraphics[width=\linewidth,trim={5cm 0 4.3cm 0},clip]{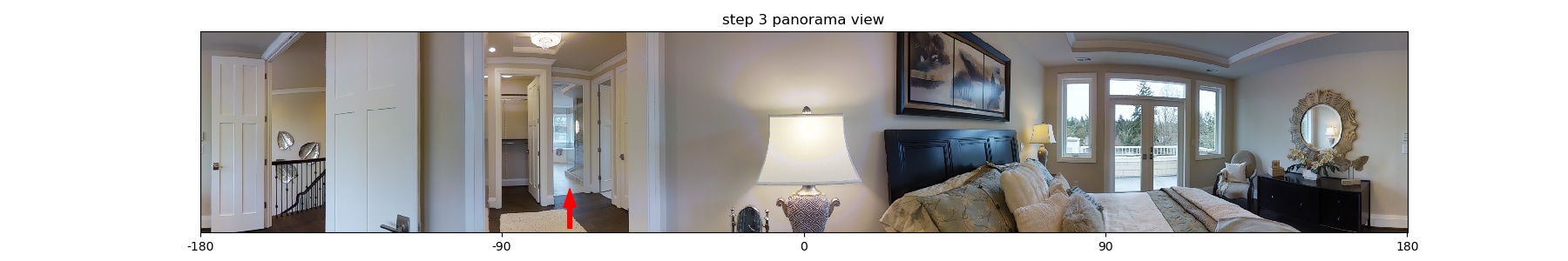}
\includegraphics[width=\linewidth,trim={5cm 0 4.3cm 0},clip]{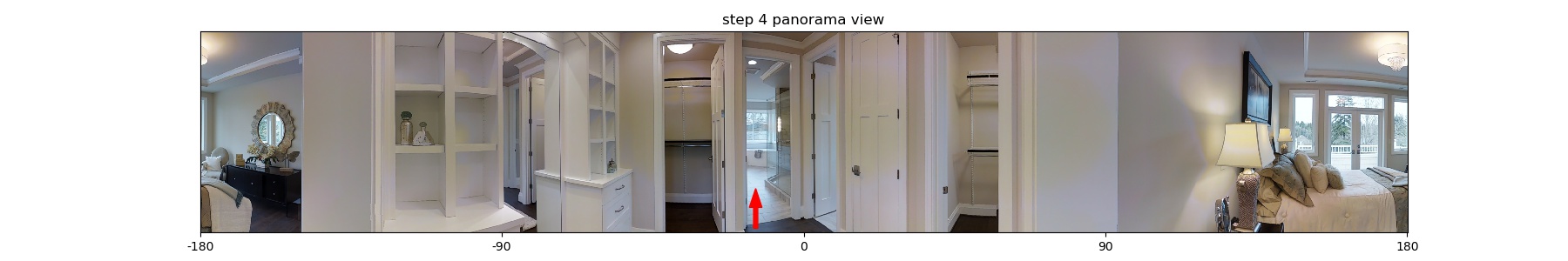}
\includegraphics[width=\linewidth,trim={5cm 0 4.3cm 0},clip]{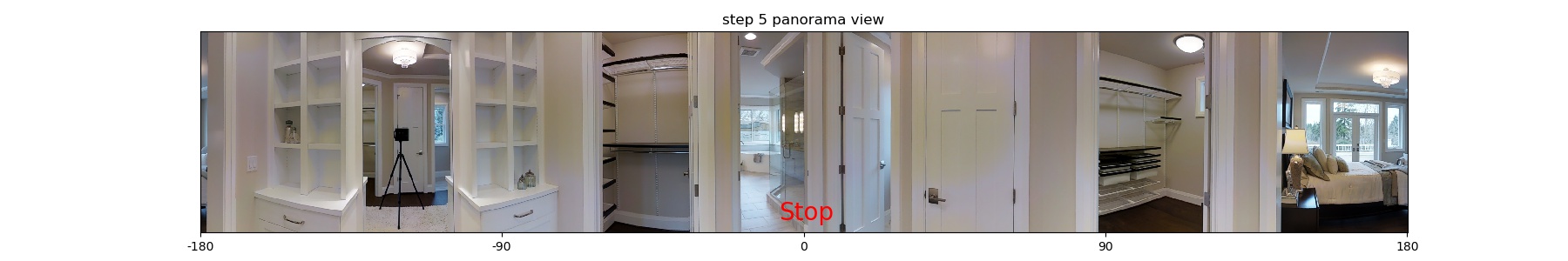}
\\ \small{Navigation steps of the panorama agent. The \textcolor{red}{red} arrow shows the direction chosen by the agent to go next.} \\
\end{center}
\caption{Follower \textbf{with pragmatic inference} on val unseen. The speaker model helps resolve the ambiguous ``walk into bedroom'' command (there are two bedrooms), allowing the follower to enter the correct bedroom on the right, where it could see a ``table clock''.}
\label{fig:supp_rational_unseen1}
\vspace{-2em}
\end{figure}

\begin{figure}[t]
\vspace{-2em}
\begin{center}
\textbf{Instruction}:\\
\textit{Enter the bedroom and make a slight right. Walk across the room near the foot of the bed. Turn right at the end of the rug. Wait near the mirror.} \\ ~ \\
\small{\textit{rear}: -180 degree~~~~~~~~~~~~\textit{left}: -90 degree~~~~~~~~~~~~~~\textit{front}: 0 degree~~~~~~~~~~\textit{right}: +90 degree~~~~~~~~~\textit{rear}: +180 degree} \\
\includegraphics[width=\linewidth,trim={5cm 0 4.3cm 0},clip]{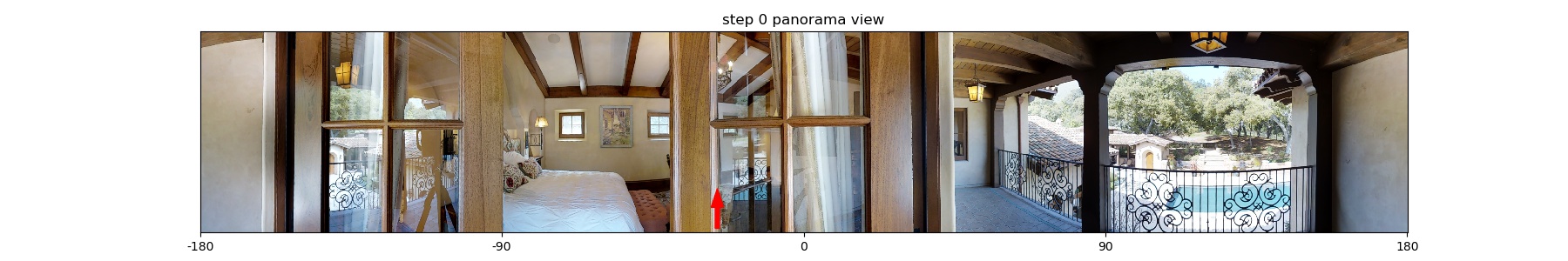}
\includegraphics[width=\linewidth,trim={5cm 0 4.3cm 0},clip]{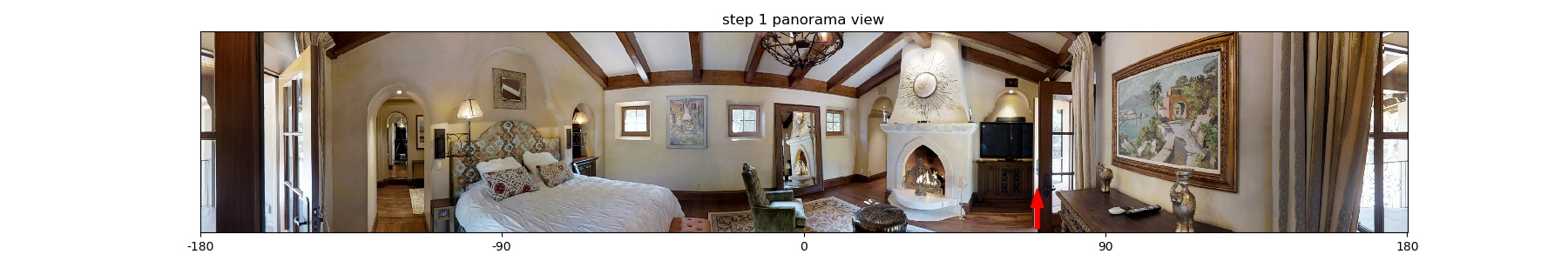}
\includegraphics[width=\linewidth,trim={5cm 0 4.3cm 0},clip]{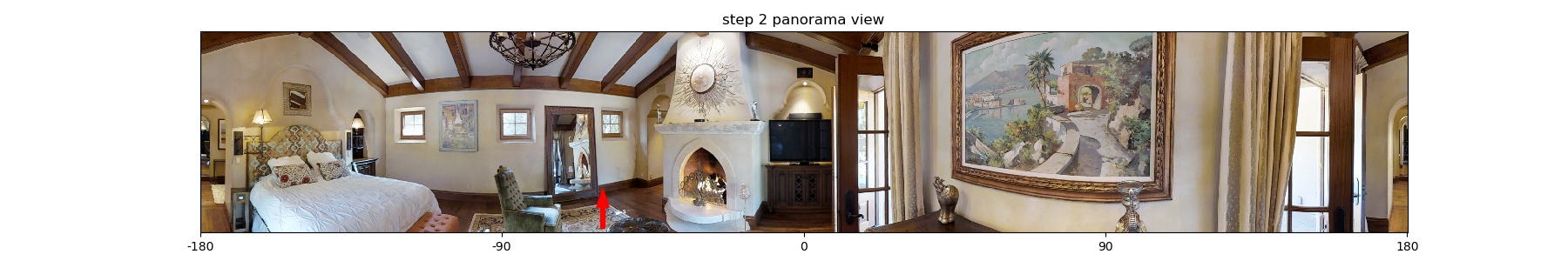}
\includegraphics[width=\linewidth,trim={5cm 0 4.3cm 0},clip]{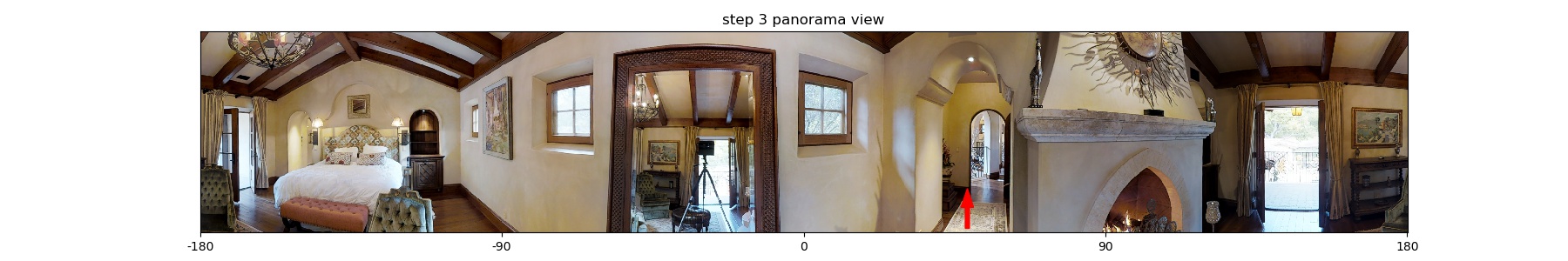}
\includegraphics[width=\linewidth,trim={5cm 0 4.3cm 0},clip]{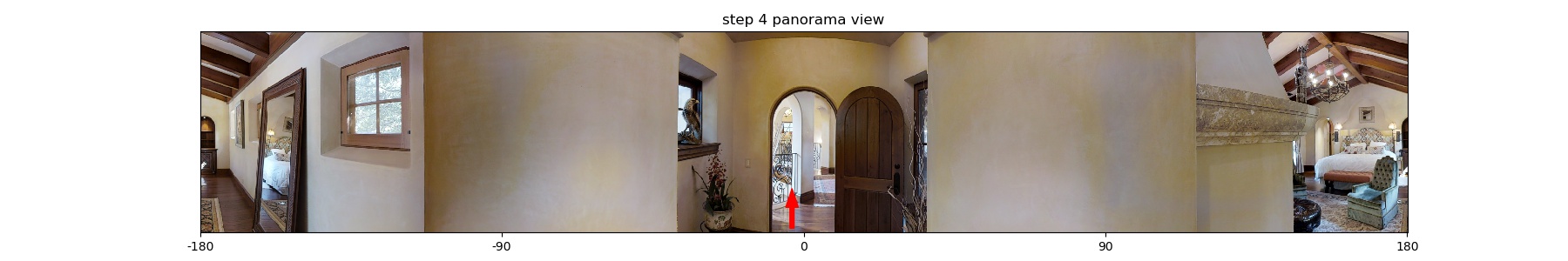}
\includegraphics[width=\linewidth,trim={5cm 0 4.3cm 0},clip]{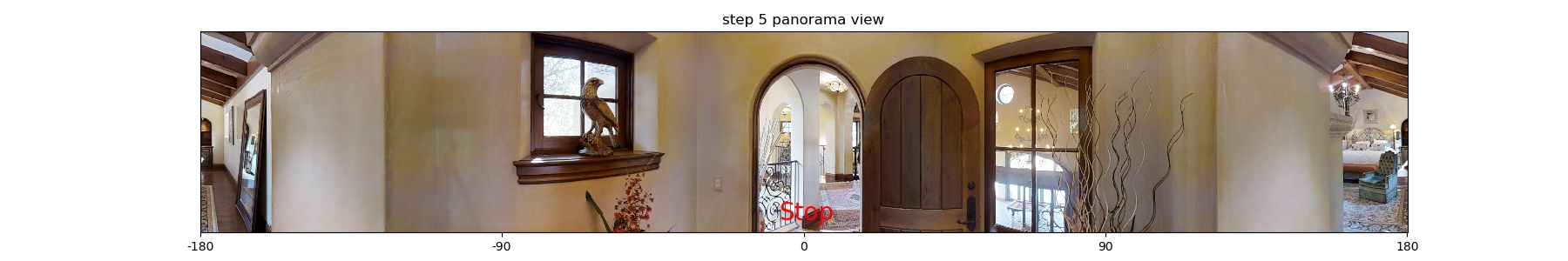}
\\ \small{Navigation steps of the panorama agent. The \textcolor{red}{red} arrow shows the direction chosen by the agent to go next.} \\
\end{center}
\caption{Follower \textbf{without pragmatic inference} on val unseen. Although making a right turn as described, the follower fails to turn right at the correct location, and stopped at the door instead of the mirror. The route taken by the follower would be better described as \textit{``...wait near the door''} by a human, which the speaker could learn to capture.}
\label{fig:supp_base_unseen2}
\vspace{-2em}
\end{figure}

\begin{figure}[t]
\vspace{-2em}
\begin{center}
\textbf{Instruction}:\\
\textit{Enter the bedroom and make a slight right. Walk across the room near the foot of the bed. Turn right at the end of the rug. Wait near the mirror.} \\ ~ \\
\small{\textit{rear}: -180 degree~~~~~~~~~~~~\textit{left}: -90 degree~~~~~~~~~~~~~~\textit{front}: 0 degree~~~~~~~~~~\textit{right}: +90 degree~~~~~~~~~\textit{rear}: +180 degree} \\
\includegraphics[width=\linewidth,trim={5cm 0 4.3cm 0},clip]{figures/398_6954_0/0.jpg}
\includegraphics[width=\linewidth,trim={5cm 0 4.3cm 0},clip]{figures/398_6954_0/1.jpg}
\includegraphics[width=\linewidth,trim={5cm 0 4.3cm 0},clip]{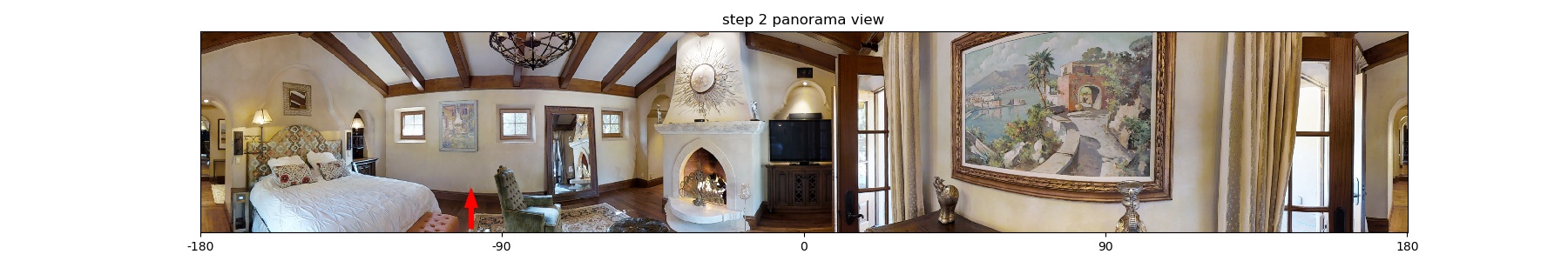}
\includegraphics[width=\linewidth,trim={5cm 0 4.3cm 0},clip]{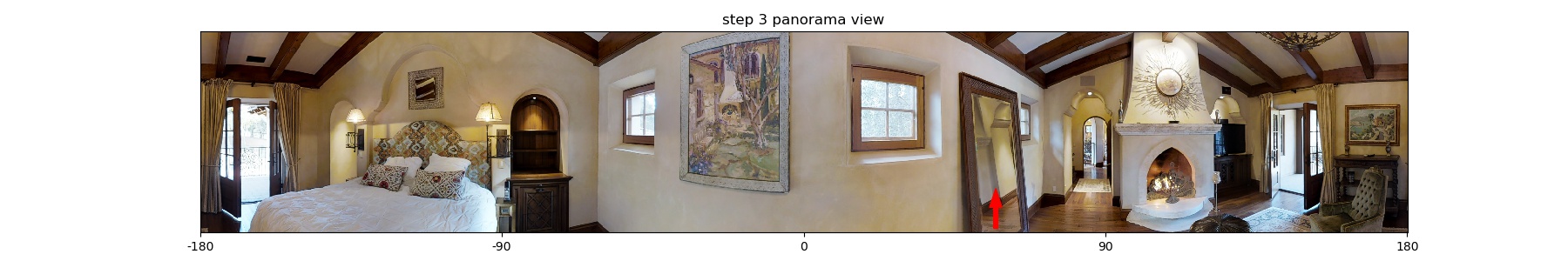}
\includegraphics[width=\linewidth,trim={5cm 0 4.3cm 0},clip]{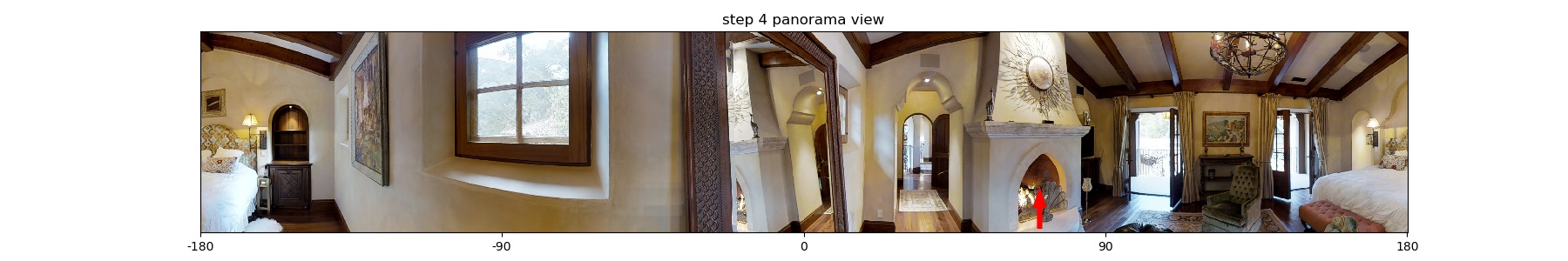}
\includegraphics[width=\linewidth,trim={5cm 0 4.3cm 0},clip]{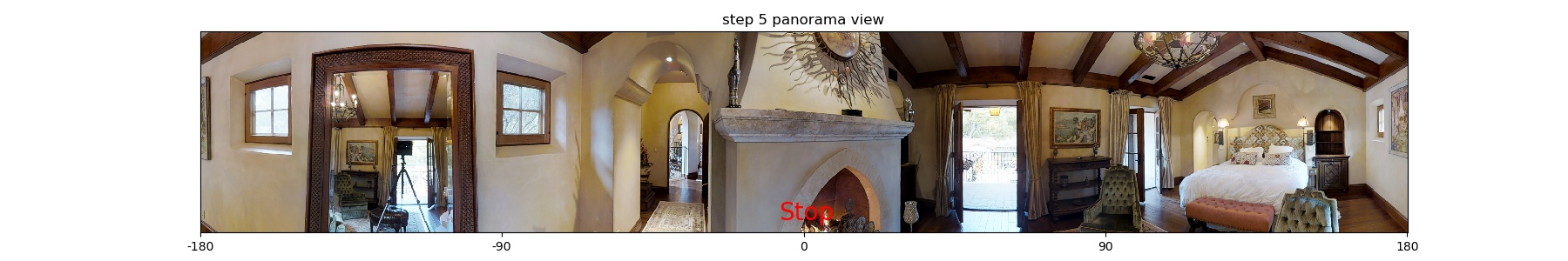}
\\ \small{Navigation steps of the panorama agent. The \textcolor{red}{red} arrow shows the direction chosen by the agent to go next.} \\
\end{center}
\caption{Follower \textbf{with pragmatic inference} on val unseen. Using the speaker to measure how likely a route matches the provided description, the follower made the right turn at the correct location \textit{``the end of the rug''}, and stopped near the mirror.}
\label{fig:supp_rational_unseen2}
\vspace{-2em}
\end{figure}

We show more examples of how the speaker model helps instruction following on both seen and unseen validation environments. Figure \ref{fig:supp_base_seen} v.s. \ref{fig:supp_rational_seen} show the step-wise navigation trajectory of the base follower (without pragmatic inference) and the follower model with pragmatic inference, on the val seen split. Figure~\ref{fig:supp_base_unseen1} v.s. Figure~\ref{fig:supp_rational_unseen1} and Figure~\ref{fig:supp_base_unseen2} v.s. Figure~\ref{fig:supp_rational_unseen2} show the trajectory of the agent without and with pragmatic inference (using the speaker model) on the val unseen split. The speaker helps disambiguate vague instructions by globally measuring how likely a route can be described by the instruction.

\begin{figure}[t]
\vspace{-2em}
\begin{center}
\textbf{Instruction}: \textit{Go through the dooorway on the right, continue straightacross the hallway and into the room ahead. Stop near the shell.} \\
\includegraphics[height=0.95\textheight,trim={0 0 0 2cm},clip]{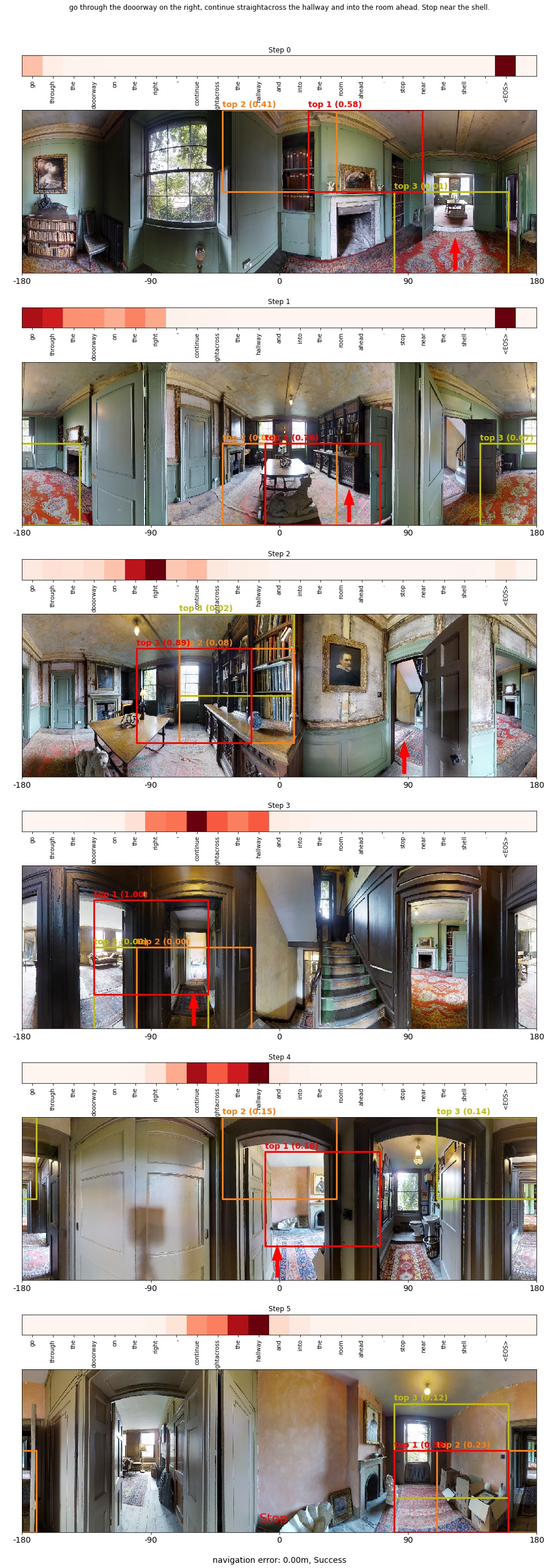} \\ 
\end{center}
\vspace{-1em}
\caption{Image and textual attention visualization on val unseen (best viewed at 200\%). At each step, the textual attention is shown at the top, and the 1st, 2nd and 3rd most attended view angles are shown in \textcolor{red}{red}, \textcolor{orange}{orange} and \textcolor{yellow}{yellow} boxes, respectively (the number in the parenthesis shows the attention weight). The red arrow shows the direction chosen by the agent to go next.}
\label{fig:supp_attention_vis1}
\vspace{-2em}
\end{figure}

\begin{figure}[t]
\vspace{-2em}
\begin{center}
\textbf{Instruction}: \textit{Walk through the kitchen, enter the dining room, walk to the doorway to the right of the dining room table, wait at the glass table.} \\
\includegraphics[height=0.95\textheight,trim={0 0 0 2cm},clip]{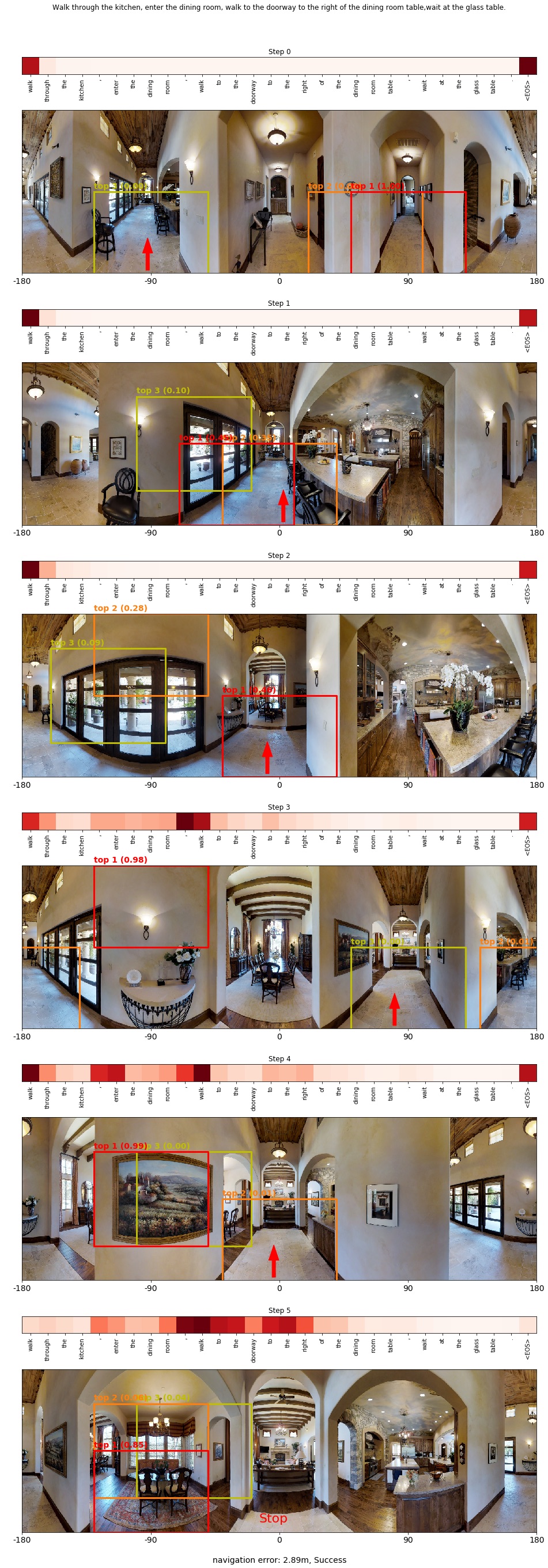} \\ 
\end{center}
\vspace{-1em}
\caption{Image and textual attention visualization on val unseen (best viewed at 200\%). At each step, the textual attention is shown at the top, and the 1st, 2nd and 3rd most attended view angles are shown in \textcolor{red}{red}, \textcolor{orange}{orange} and \textcolor{yellow}{yellow} boxes, respectively (the number in the parenthesis shows the attention weight). The red arrow shows the direction chosen by the agent to go next.}
\label{fig:supp_attention_vis2}
\vspace{-2em}
\end{figure}

\begin{figure}[t]
\vspace{-2em}
\begin{center}
\textbf{Instruction}: \textit{Walk up stairs. Turn left and walk to the double doors by the living room.} \\
\includegraphics[height=0.95\textheight,trim={0 0 0 2cm},clip]{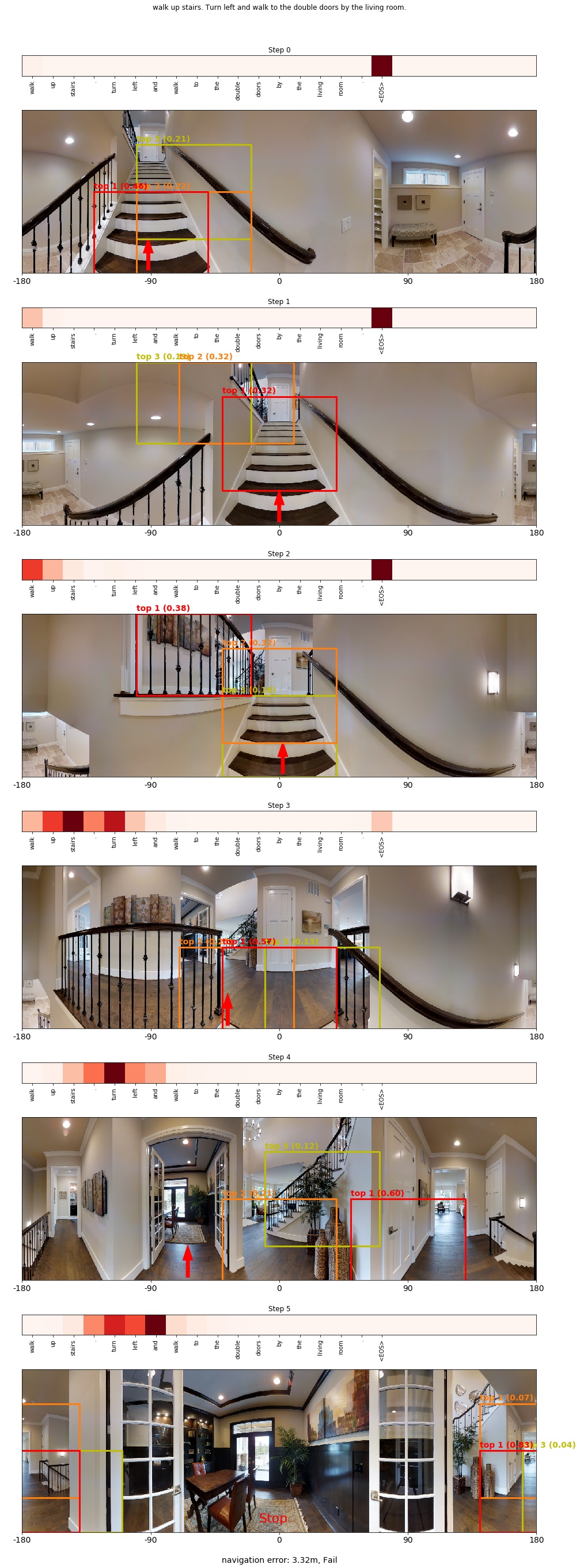} \\ 
\end{center}
\vspace{-1em}
\caption{Image and textual attention visualization on val unseen (best viewed at 200\%). At each step, the textual attention is shown at the top, and the 1st, 2nd and 3rd most attended view angles are shown in \textcolor{red}{red}, \textcolor{orange}{orange} and \textcolor{yellow}{yellow} boxes, respectively (the number in the parenthesis shows the attention weight). The red arrow shows the direction chosen by the agent to go next.}
\label{fig:supp_attention_vis3}
\vspace{-2em}
\end{figure}

We also visualize the image attention (attention weights $\alpha_{t,i}$ of each view angle $i$ in our panoramic action space in Sec.~3.3 in the main paper), and the textual attention on the input instructions from the sequence-to-sequence model in Figures~\ref{fig:supp_attention_vis1}, \ref{fig:supp_attention_vis2} and \ref{fig:supp_attention_vis3}.

\section{Submission to Vision and Language Navigation Challenge}
\label{sec:challenge}

We participated in the Vision and Language Navigation Challenge\footnote{\url{https://evalai.cloudcv.org/web/challenges/challenge-page/97/overview}}, an online challenge for the vision-and-language navigation task on the R2R dataset. We submitted the predictions from our full method to the evaluation server, using single models for the speaker and listener, without any additional ensembling. At the time of writing, our method (under the team name ``Speaker-Follower'') remains the top-performing method on the challenge leader-board with a success rate of 53.49\% (the same success rate as in the Table~2 test split in the main paper).

When generating the predictions for  the  challenge submission, we modified the method for generating  final routes to comply with the challenge guidelines. In Table~1, Table~2 in the main paper and Table~\ref{tab:supp_ablations}, the performance of our full model with pragmatic inference is evaluated using a single top-ranking candidate route, where the candidate routes are generated with state-factored search in Sec.~\ref{sec:supp_state_factored_search}.
Hence, our reported navigation error, success rate and oracle success rate are all computed by  choosing a single candidate route per instruction.
However, the challenge guidelines require that the submitted trajectories must be generated from \textit{a single independent evaluation run}, where the agent must move sequentially and all its movements during inference must be recorded. So just returning the route selected by pragmatic inference, or by search, would violate the contest guidelines, as this route may not contain all world states explored during search.
On the other hand, the agent is allowed to backtrack to a previous location on its path, as long as the agent does not teleport and all its movements are recorded in the final trajectory (which we confirmed with the challenge organizer).

To comply with the guidelines and maintain physical plausibility, we log all states visited by the search algorithm in the order they are traversed.
The agent expands each route one step forward at a time, and then switches to expand the next route. When switching from one route to another route, the agent first backtracks to the closest common ancestor state of the two routes in  the search (which could be the starting location). From there it goes to the frontier of the other route and takes an action there to expand it. Once the set of candidate routes has been completed, they are ranked according to Equation~1 in the main paper for pragmatic inference, selecting a route that ends at a target location. Finally, the agent moves from its last visited location in the search to this target location. It then takes the stop action to end the episode. As a result of this, all the agent's movements, including route expansion and route switching, are recorded in its final trajectory.

By design, the sequential inference procedure above yields \textit{exactly the same success rate} as the pragmatic inference procedure, since it returns routes with the same end states. Unsurprisingly, the oracle success rate increases substantially (from $63.9\%$ to $96.0\%$), since the resulting final trajectory records the visited locations from all routes and the oracle success rate is determined by distance between the ground-truth target location and the closest visited location. We also note that the agent's final trajectory is very long ($1257.38\mathrm{m}$ per instruction on average) since it needs to walk a substantial distance to sequentially expand all candidate routes and to switch between them.

In addition, we also evaluated the predictions from our model without pragmatic inference on the challenge server. Without pragmatic inference, our model achieves 35.08\% success rate, 44.45\% oracle success rate and $6.62\mathrm{m}$ navigation error on the challenge test set, with an average trajectory length of $14.82\mathrm{m}$. This is close to the performance of our model under the same setting on val unseen split (Row 6 in Table~1 in the main paper).

Finally, we note that our method in this work is designed mostly to optimize success rate and navigation error, and we leave potential improvement to reduce trajectory length via inference (such as ordering the routes by location to reduce switching overhead) and modeling (such as building speaker models that can rank partial incomplete routes) to future work.

\end{document}